\pdfoutput=1

\documentclass[11pt]{article}

\usepackage[preprint]{acl}

\usepackage{times}
\usepackage{latexsym}
\usepackage{graphicx} 
\usepackage{booktabs} 
\usepackage{multirow} 
\usepackage{amsmath}  

\usepackage[T1]{fontenc}
\usepackage{multirow}
\usepackage[utf8]{inputenc}

\usepackage{microtype}
\usepackage{makecell}

\usepackage{inconsolata}

\usepackage{graphicx}
\usepackage{amsmath}
\usepackage{amsfonts} 

\usepackage[utf8]{inputenc}
\usepackage{amsmath} 
\usepackage{amssymb} 
\usepackage{algorithm} 
\usepackage{algpseudocode} 
\usepackage{booktabs} 
\usepackage{graphicx} 
\usepackage{hyperref} 
\usepackage{enumitem}
\usepackage{makecell}

\usepackage{booktabs} 

\usepackage{multirow}
\usepackage{colortbl}
\usepackage[normalem]{ulem}
\useunder{\uline}{\ul}{}

\title{GRASP: Replace Redundant Layers with Adaptive Singular Parameters for Efficient Model Compression}

\author{
  \textbf{Kainan Liu}\textsuperscript{1,2,\textdagger}, 
  \textbf{Yong Zhang}\textsuperscript{1,\textdagger},
  \textbf{Ning Cheng}\textsuperscript{1,\textasteriskcentered},
  \textbf{Zhitao Li}\textsuperscript{1}, \\
  \textbf{Shaojun Wang}\textsuperscript{1},
  \textbf{Jing Xiao}\textsuperscript{1},  \\
  \textsuperscript{1}Ping An Technology (Shenzhen) Co., Ltd., China \\
  \textsuperscript{2}The Hong Kong University of Science and Technology (Guangzhou) \\
  \texttt{\{zhangyong203, chengning211\}@pingan.com.cn}
}

\begin{document}
\maketitle

\setlength{\footnotesep}{0.4\baselineskip} 

\renewcommand{\thefootnote}{}\setcounter{footnote}{0}
\footnotetext{\textdagger\ Equal contribution.}
\footnotetext{\textasteriskcentered \ Corresponding author.}\setcounter{footnote}{0}
\footnotetext{This work was done during Kainan Liu’s internship at Ping An Technology.}\setcounter{footnote}{0}

\renewcommand{\thefootnote}{\arabic{footnote}}

\begin{abstract}
Recent studies have demonstrated that many layers are functionally redundant in large language models (LLMs), enabling model compression by removing these layers to reduce inference cost. While such approaches can improve efficiency, indiscriminate layer pruning often results in significant performance degradation. In this paper, we propose \textbf{GRASP} (\underline{G}radient-based \underline{R}etention of \underline{A}daptive \underline{S}ingular \underline{P}arameters), a novel compression framework that mitigates this issue by preserving sensitivity-aware singular values. Unlike direct layer pruning, GRASP leverages gradient-based attribution on a small calibration dataset to adaptively identify and retain critical singular components. By replacing redundant layers with only a minimal set of parameters, GRASP achieves efficient compression while maintaining strong performance with minimal overhead. Experiments across multiple LLMs show that GRASP consistently outperforms existing compression methods, achieving 90\% of the original model's performance under 20\% compression ratio. The source code is available at \href{https://github.com/LyoAI/GRASP}{https://github.com/LyoAI/GRASP}.

\end{abstract}

\section{Introduction}
Large Language Models (LLMs) have demonstrated remarkable capabilities across a wide range of tasks, including language generation, reasoning, and question answering (\citealp{NEURIPS2020_1457c0d6}; \citealp{touvron2023llama}). However, their massive parameter sizes pose computational and memory challenges, hindering deployment on resource-limited devices (\citealp{zhou2024survey}). To address this, model compression techniques such as quantization (\citealp{frantar2022gptq}; \citealp{lin2024awq}; \citealp{xiao2023smoothquant}), knowledge distillation (\citealp{gu2023knowledge}; \citealp{xu2024survey}), and pruning(\citealp{sun2024a}; \citealp{ashkboos2024slicegpt}) have been widely explored. Among these, structured pruning methods remove entire components such as neurons or layers to streamline the model, thereby achieving hardware efficiency and inference speedup.

\begin{figure}[tp]
    \centering
    \vspace{-1mm}
    \includegraphics[width=1\linewidth]{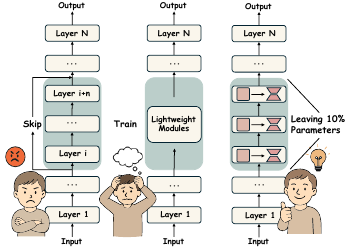}
\caption{Unlike conventional layer pruning, which either skips redundant layers—often causing moderate performance drops—or replaces them with lightweight modules that require additional training, GRASP (right) retains only the most critical 10\% of parameters within the redundant layers, effectively preserving accuracy with minimal overhead.}
    \label{fig:Main}
    \vspace{-4mm}
\end{figure}

In this work, we focus on structured layer pruning, which builds on prior findings that certain consecutive layers in large language models (LLMs) are functionally redundant. These findings have inspired approaches that either remove such layers entirely~(\citealp{men2024shortgpt}, \citealp{yang2024laco}, \citealp{kim2024shortened}) or replace them with lightweight modules~(\citealp{chen2024compressing}) to reduce inference cost. Although layer removal is simple and computationally efficient, it often results in significant performance degradation. This degradation arises from disrupted information flow and misaligned intermediate representations~(\citealp{liang2024internal}), suggesting that layer removal eliminates certain important components that contribute meaningfully to maintaining model performance. Replacing redundant layers with lightweight modules can mitigate this issue to some extent, but such modules are typically randomly initialized, requiring substantial computational resources for training.

In this paper, we propose \textbf{GRASP} (\underline{G}radient-based \underline{R}etention of \underline{A}daptive \underline{S}ingular \underline{P}arameters), a novel compression framework that replaces redundant layers with adaptive singular parameters for efficient LLM compression. Unlike direct layer removal, GRASP exploits the low-rank structure inherent in redundant layers, replacing the redundant layers with only a small subset of parameters while maintaining strong model performance. Specifically, GRASP operates in two key steps: First, it identifies layers suitable for pruning based on the cosine similarity of output hidden states between adjacent layers. Then, instead of relying on magnitude-based heuristics, GRASP leverages gradient attribution derived from a small calibration dataset to adaptively identify and retain the singular values most critical for downstream task performance.

To evaluate the effectiveness of GRASP, We conduct extensive experiments on 19 datasets and 5 models from two distinct LLM families (LLaMA and Mistral). Notably, GRASP achieves strong performance in a training-free setting, requiring no additional optimization. Furthermore, when post-training compensation is applied, only a small number of samples are needed to rapidly restore model performance. This efficiency arises from retaining critical components within redundant layers, rather than relying on randomly initialized replacements. Overall, this paper makes the following contributions:

\begin{itemize}[leftmargin=10pt, itemindent=0pt]
\item We propose \textbf{GRASP}, a novel training-free compression framework that replaces redundant layers with adaptive singular parameters, leveraging the low-rank structure within LLMs to preserve performance with minimal overhead.
\item We introduce a gradient-based singular value selection mechanism, enabling efficient identification of critical components without relying on magnitude-based heuristics.
\item We conduct extensive experiments across ten datasets and five models from two major LLM families (LLaMA and Mistral), demonstrating that GRASP consistently achieves strong performance under both training-free and low-resource fine-tuning settings.
\end{itemize}

\section{Method}
\label{sec:method}
Figure~\ref{fig:Main} illustrates the workflow of GRASP. The method consists of two main steps: Identifying redundant layers to be pruned (Sec~\ref{subsec:redundant_layer_selection}) and replacing redundant layers with critical singular components guided by gradient-based attribution (Sec~\ref{subsec:Gradient-Guided Singular Value Selection}). Below, we describe each step in detail.

\subsection{Redundant Layer Selection}
\label{subsec:redundant_layer_selection}
The first step in GRASP is identifying redundant layers. These are layers that contribute minimally to the transformation of hidden states, exhibiting high redundancy and limited impact on overall model performance. Following prior works (\citealp{song2024sleb}; \citealp{chen2024compressing}), we use cosine similarity to quantify the degree of transformation in a given layer.

For a transformer layer with input hidden state \(H_i \in \mathbb{R}^d\) and output hidden state \(H_{i+1} \in \mathbb{R}^d\), the cosine similarity is computed as:
\begin{equation}
\label{eq:cosine_similarity}
\text{cos}(H_i, H_{i+1}) = \frac{H_i^T H_{i+1}}{\|H_i\|_2 \|H_{i+1}\|_2}
\end{equation}
A high cosine similarity indicates minimal transformation, suggesting that the layer is redundant. Instead of directly removing these layers, GRASP compresses weight matrices with a gradient-based approach to retain critical internal transformations.

\subsection{Layer Replacement with Adaptive Singular Parameters}
\label{subsec:Gradient-Guided Singular Value Selection}
\begin{figure}[tp]
    \centering
    \includegraphics[width=1\linewidth]{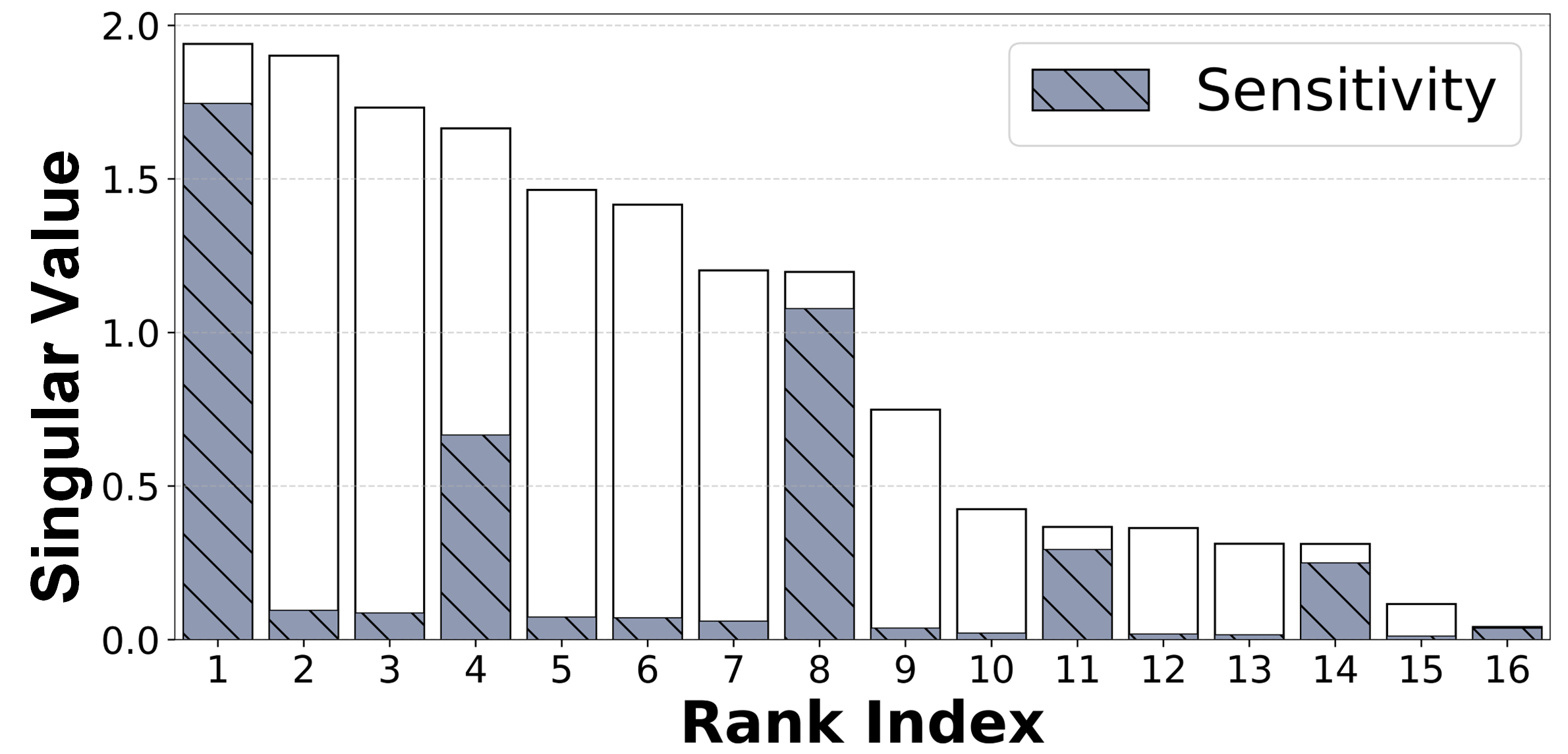}
    \caption{Sensitivity analysis of grouped singular value truncation. While singular values are typically ordered by magnitude, their impact on downstream performance does not follow the same order.}
    \label{fig:Method}
    \vspace{-5mm}
\end{figure}

\paragraph{Motivation.} GRASP operates under the hypothesis that redundant layers exhibit an inherent low-rank structure, allowing their functionality to be effectively approximated using low-rank matrices. Based on this insight, a straightforward approach is to apply singular value decomposition (SVD) to these layers. However, prioritizing components solely based on singular value magnitude does not necessarily correlate with downstream task performance~(\citealp{hsu2022language, hua2025dynamic}). To validate this point, we selectively zero out groups of singular values in the weight matrices of large language models and measure their impact on downstream tasks. As shown in Figure~\ref{fig:Method}, we make two key observations: (1) The contribution of a singular value to downstream task performance is not solely determined by its magnitude; smaller values can be crucial for task performance in some cases. (2) Redundant layers exhibit a highly low-rank structure, where only a few singular directions dominate the model performance. 

\paragraph{Key Design.} To address this limitation, GRASP introduces gradient-based attribution to evaluate the importance of each singular value by its contribution to model performance, rather than relying on magnitude alone. Formally, the singular value decomposition of a weight matrix \( W \in \mathbb{R}^{m \times n} \) is given by \(W = U \Sigma V^{\top}\). Equivalently, \( W \) can be expressed as the sum of its rank-one components:
\begin{equation}
    W = \sum_{k=1}^{l} u_k \sigma_k v_k^{\top},
\end{equation}
where \( l = \min(m, n) \), \( \sigma_k \) denotes the \(k\)-th singular value, and \( u_k \), \( v_k \) are the \(k\)-th column vectors of the orthogonal matrices \( U \) and \( V \), respectively. For brevity, we use \(\Phi_k\) to represent the singular group \(\{u_k, \sigma_k, v_k^{\top}\}\). GRASP estimates the importance of each \(\Phi_k\) using a small, general-purpose calibration dataset (e.g., WikiText-2), computing a sensitivity-based score that reflects its effect on model performance, given by:
\begin{equation}
    I(\Phi_k) = T(\sigma_k) + \sum_{i=1}^{m}T(u_{k, i}) + \sum_{i=1}^{n}T(v_{k, i})
\end{equation}
where \(T(\cdot)\) denotes the estimated loss change when a certain parameter \(\theta\) is zeroed out. This can be approximated by the second-order Taylor expansion~(\citealp{lecun1989optimal}), defined as:
\begin{equation}
    T(\theta) = \left| \theta^{\top} \nabla_{\theta} \mathcal{L} + \frac{1}{2} \theta^{\top} H \theta + \mathcal{O}(\|\theta\|^{3}) \right|
\end{equation}
\begin{equation}
\mathcal{L} = - \sum_{t} \log P(y_t \mid x_{\leq t})
\end{equation}
where \(\nabla_{\theta} \mathcal{L}\) denotes the gradient of the standard language modeling objective function \(\mathcal{L}\) w.r.t. the parameter \(\theta\), and \(H\) represents the corresponding Hessian matrix. To reduce computational overhead, we omit the second-order term and approximate the importance using only the first-order derivative~(\citealp{hua2025dynamic}, \citealp{kim2024shortened}):
{\small
\begin{equation}
\label{Eq:Taylor-Oriented Decomposition}
I(\Phi_k) = \left| \sigma_k \frac{\partial \mathcal{L}}{\partial \sigma_k} \right| 
+ \sum_{i=1}^{m} \left| u_{k,i} \frac{\partial \mathcal{L}}{\partial u_{k,i}} \right|
+ \sum_{i=1}^{n} \left| v_{k,i} \frac{\partial \mathcal{L}}{\partial v_{k,i}} \right|
\end{equation}
}
The first term in Eq.~\ref{Eq:Taylor-Oriented Decomposition} captures the gradient projection of the loss \(\mathcal{L}\) w.r.t. the singular value \(\sigma_i\). The proof is given in Appendix\ref{appendix:Proof}. By aggregating the first-order expansions across all components within each singular group, our method effectively captures the contribution of singular values, where larger values indicate greater importance. This gradient-oriented attribution moves beyond heuristic magnitude-based criteria, enabling performance-aligned importance evaluation. Under the hypothesis that redundant layers exhibit an inherent low-rank structure, we retain only the top-$r$\% singular values most critical to model performance based on Eq.~\ref{Eq:Taylor-Oriented Decomposition}, and use them to replace the corresponding redundant layer. The relationship between the singular group retain ratio ($r$\%) and the overall target compression ratio is detailed in Appendix~\ref{appendix:retain_ratio}.

\subsection{Detailed Implementation of GRASP}
\label{method:Layer-wise Low-Rank Decomposition}
GRASP processes the redundant layers in a sequential manner, starting from the final redundant layer and proceeding backward through the network. We summarize the detailed algorithm in Algorithm~\ref{alg:GRASP}.

\begin{algorithm}[H]
\caption{GRASP: Gradient-based Retention of Adaptive Singular Parameters}
\label{alg:GRASP}
\begin{algorithmic}[1]
\Require Model $M$, Calibration set $D$, Retain ratio $r$
\Ensure Compressed model $\tilde{M}$
\vspace{2pt}
\State \textbf{Step 1: Redundant Layer Selection:}
\State Compute cosine similarity $\text{cos}(H_i, H_{i+1})$ for all layers via Eq.\ref{eq:cosine_similarity} using \(D\)
\State Select top-$L$ layers with highest similarity as redundant

\vspace{2pt}
\State \textbf{Step 2: Gradient-Guided Compression:}
\For{each redundant layer $l$ (in reverse order)}
    \For{each $W \in \{$attention, MLP$\}$}
        \State SVD: $W = U \Sigma V^\top$
        \State Compute importance $I(\Phi_k)$ for each $\Phi_k = \{u_k, \sigma_k, v_k^\top\}$ via Eq.\ref{Eq:Taylor-Oriented Decomposition} using \(D\)
        \State Keep top-$r\%$ singular groups, reconstruct $\tilde{W}$
    \EndFor
\EndFor
\State \Return $\tilde{M}$
\end{algorithmic}
\end{algorithm}

\section{Experiments}

In this section, we conduct comprehensive experiments to evaluate GRASP from three key perspectives. (1) We first compare our method with existing pruning-based LLM compression approaches to demonstrate its effectiveness (Section \ref{exp:Comparison with Pruning-based LLM Compression Methods}). (2) Next, we analyze the inference speed-up achieved by GRASP (Section \ref{exp:Inference Efficiency of GRASP}). (3) Finally, we investigate the factors influencing our approach by performing ablation studies on the choice of calibration datasets and pruning strategies. (Section~\ref{sec:ablation_study})

\subsection{Experimental Setup}
Below we detail the models, benchmarks, baselines and implementation details used in our experiments, with more experimental setups provided in Appendix \ref{appendix:Experimental Setup and Hyperparameters Configuration}.

\paragraph{Models.} We evaluate GRASP on a range of large language models (LLMs) from two model families: the LLaMA family, including LLaMA-7B~(\citealp{touvron2023llama1}), LLaMA 2-7B, LLaMA 2-13B~(\citealp{touvron2023llama}), and LLaMA 3.1-8B-Instruct~(\citealp{dubey2024llama}), as well as Mistral-7B~(\citealp{jiang2023mistral}) from the Mistral family.

\paragraph{Baselines.} We compare GRASP against 8 structured pruning methods to substantiate its efficacy:
\begin{itemize}[leftmargin=10pt, itemindent=2pt]
\item \textbf{Layer-pruning methods} We consider three representative layer-pruning methods as baselines: ShortGPT (\citealp{men2024shortgpt}), LaCo~(\citealp{yang2024laco}) and LLM-Streamline~(\citealp{chen2024compressing}).
\item \textbf{Module-pruning methods} We also select LLM-Pruner~(\citealp{ma2023llmpruner}) and SliceGPT~(\citealp{ashkboos2024slicegpt}) which prune the redundant modules in LLMs.
\item \textbf{Low-rank Pruning methods} Considering our method involves Gradient-based SVD, we also compare with other low-rank pruning methods: FWSVD~(\citealp{hsu2022language}), ASVD~(\citealp{yuan2023asvd}) and SVD-LLM~(\citealp{wang2024svd}).
\end{itemize}
We provide a detailed comparison of these pruning-based LLM compression methods in Appendix \ref{appendix:Comparison of Concurrent Pruning-based Methods}.

\paragraph{Implementation Details.} To ensure a fair comparison, we randomly sample 512 data points from the \textbf{WikiText-2} dataset as the calibration dataset. All experiments are conducted on NVIDIA A100-SXM4 (80GB) GPUs. Further experimental details can be found in Appendix \ref{appendix:Experimental Setup and Hyperparameters Configuration}.

\begin{table*}[t]
\centering
\resizebox{0.85\linewidth}{!}{%
    \begin{tabular}{c|ccccccc|cc}
    \toprule[2pt]
    \textbf{Methods} & \textbf{OpenbookQA} & \textbf{ARC\_e} & \textbf{WinoGrande} & \textbf{HellaSwag} & \textbf{ARC\_c} & \textbf{PIQA} & \textbf{MathQA} & \textbf{Average}  & \textbf{Percentage} \\
    \midrule
    Dense & 0.34  & 0.82  & 0.74  & 0.59  & 0.52  & 0.80  & 0.39  & 0.60  & 100.0\% \\
    \midrule
    LaCo  & \textbf{0.26}  & 0.49  & 0.65  & 0.33  & 0.30  & 0.65  & \textbf{0.30}  & 0.42  & 70.9\% \\
    ShortGPT & 0.21  & 0.57  & 0.66  & 0.42  & 0.32  & 0.67  & 0.26  & 0.44  & 74.1\% \\
    SliceGPT & 0.15  & 0.43  & 0.51  & 0.30  & 0.23  & 0.58  & 0.22  & 0.35  & 57.7\% \\
    GRASP & 0.22  & \textbf{0.60}  & \textbf{0.70}  & \textbf{0.44}  & \textbf{0.37} & \textbf{0.69} & 0.28  & \textbf{0.47}  & \textbf{78.6\%} \\
    \bottomrule[2pt]
\end{tabular}%
}
\caption{Zero-shot performance of GRASP and structured pruning baselines without post-training under a 20\% compression ratio. Results are reported on seven reasoning datasets (individual and average accuracy). Bold values indicate the best performance.}
\label{tab:Comparison with pruning-based methods without compensation}
\end{table*}

\begin{table*}[t]
\centering
\resizebox{0.96\linewidth}{!}{%
\begin{tabular}{c|cccccccccccc|cc}
    \toprule[2pt]
    \textbf{Method} & \textbf{C3}  & \textbf{CMNLI} & \textbf{CHID} & \textbf{BoolQ} & \textbf{WSC} & \textbf{CoQA}  & \textbf{HeSW} & \textbf{PIQA} & \textbf{Race-M} & \textbf{Race-H} & \textbf{MMLU} & \textbf{CMMLU} & \textbf{Avg.} & \textbf{Per.} \\
    \midrule
    Dense & 43.8  & 33.0  & 41.6  & 70.8  & 37.5  & 66.7  & 71.3  & 78.1  & 33.1  & 35.5  & 46.8  & 31.8  & 49.2  & 100.0\% \\
    \midrule
    LLMPruner* & 29.7  & 33.4  & \textbf{28.4} & 58.7  & 40.4  & 48.5  & 54.6  & 72.0  & 22.9  & 22    & 25.3  & 25.0  & 38.4  & 78.1\% \\
    SliceGPT* & 31.5  & 31.6  & 18.5  & 59.9  & \textbf{43.3} & 49.6  & 47.5  & 68.3  & 27.0  & 29.4  & 28.8  & 24.8  & 38.4  & 78.0\% \\
    LaCo* & 39.7  & 34.4  & 36.1  & 64.1  & 40.4  & 45.7  & 55.7  & 69.8  & 23.6  & 22.6  & 26.5  & 25.2  & 40.3  & 82.0\% \\
    ShortGPT* & 40.2  & 34.4  & 21.5  & 67.3  & 40.4  & 51.7  & 59.7  & 69.0  & 35.2  & 34.7  & 44.6  & 28.9  & 44.0  & 89.4\% \\
    LLM-Streamline-FFN* & 40.7  & 33.0  & 22.8  & 65.9  & 38.5  & 60.6  & 61.2  & 71.2  & \textbf{38.0} & \textbf{38.7} & \textbf{47.0} & \textbf{31.7} & 45.8  & 93.1\% \\
    LLM-Streamline-Layer* & 43.3  & 33.0  & 24.1  & 67.5  & 36.5  & 59.2  & 61.1  & 71.5  & 34.8  & 37.0  & 45.5  & 29.4  & 45.2  & 92.0\% \\
    \midrule
    GRASP & \textbf{44.6} & \textbf{35.1} & 26.2  & \textbf{68.4} & 41.4  & \textbf{63.2} & \textbf{62.7} & \textbf{73.3} & 35.1  & 36.1  & 43.1  & 30.7  & \textbf{46.7} & \textbf{94.9\%} \\
    \bottomrule[2pt]
\end{tabular}%
}
\caption{Comparison between GRASP and structured pruning baselines with post-training compensation under a 25\% compression ratio. Results marked with * are reported from~\citep{chen2024compressing}. Bold values indicate the best performance.}
\label{tab:Comparison with pruning-based methods with compensation}
\end{table*}

\begin{table*}[t]
\centering
\resizebox{0.9\linewidth}{!}{%
\begin{tabular}{c|c|cccccccc|cc}
\toprule[2pt]
Ratio & Method & Openb. & ARC\_e & WinoG. & HeSW & ARC\_c & PIQA  & MathQA & GSM8K & Average & Percentage \\
\midrule
0\%   & Original & 0.28  & 0.67  & 0.67  & 0.56  & 0.38  & 0.78  & 0.27  & 0.09  & 0.46 & 100.0 \\
\midrule
\multirow{5}[4]{*}{20\%} & FWSVD & 0.15  & 0.31  & 0.50  & 0.26  & 0.23  & 0.56  & 0.21  & 0.00  & 0.28 & 60.8 \\
      & ASVD & \textbf{0.25} & 0.53  & 0.64  & 0.41  & 0.27  & 0.68  & 0.24  & 0.04  & 0.38 & 82.6 \\
      & SVD-LLM & 0.22  & 0.58  & 0.63  & 0.43  & 0.29  & 0.69  & 0.24  & \textbf{0.05} & 0.39 & 84.7 \\
\cmidrule{2-12}          & Ours & 0.22  & 0.52  & \textbf{0.64} & 0.43  & 0.32  & 0.70  & 0.24  & 0.03  & 0.39 & 84.7 \\
      & Ours* & 0.24  & \textbf{0.59} & 0.63  & \textbf{0.5} & \textbf{0.35} & \textbf{0.73} & \textbf{0.25} & 0.04  & \textbf{0.42} & \textbf{91.3} \\
\midrule
\multirow{5}[4]{*}{30\%} & FWSVD & 0.17  & 0.26  & 0.49  & 0.26  & 0.22  & 0.51  & 0.19  & 0.00  & 0.26 & 56.5 \\
      & ASVD & 0.18  & 0.43  & 0.53  & 0.37  & 0.25  & 0.65  & 0.21  & 0.00  & 0.33 & 71.7 \\
      & SVD-LLM & 0.20  & 0.48  & 0.59  & 0.37  & 0.26  & 0.65  & 0.22  & 0.03  & 0.35 & 76.1 \\
\cmidrule{2-12}          & Ours & 0.19  & 0.42  & 0.62  & 0.39  & 0.28  & 0.64  & 0.23  & 0.02  & 0.35 & 76.1\\
      & Ours* & \textbf{0.24} & \textbf{0.54} & \textbf{0.64} & \textbf{0.46} & \textbf{0.32} & \textbf{0.69} & \textbf{0.24} & \textbf{0.04} & \textbf{0.40} & \textbf{87.0} \\
\midrule
\multirow{5}[4]{*}{40\%} & FWSVD & 0.16  & 0.26  & 0.51  & 0.26  & 0.22  & 0.53  & 0.21  & 0.00  & 0.27 & 58.7 \\
      & ASVD & 0.13  & 0.28  & 0.48  & 0.26  & 0.22  & 0.55  & 0.19  & 0.00  & 0.26 & 56.5 \\
      & SVD-LLM & 0.19  & 0.42  & 0.58  & 0.33  & 0.25  & 0.60  & 0.21  & \textbf{0.02}  & 0.33 & 71.7 \\
\cmidrule{2-12}          & Ours & 0.18  & 0.37  & 0.57  & 0.35  & 0.27  & 0.61  & 0.21  & 0.01  & 0.32 & 69.6\\
      & Ours* & \textbf{0.22} & \textbf{0.49} & \textbf{0.63} & \textbf{0.43} & \textbf{0.3} & \textbf{0.68} & \textbf{0.23} & \textbf{0.02} & \textbf{0.38} & \textbf{82.6}\\
\midrule
\multirow{5}[4]{*}{50\%} & FWSVD & 0.12  & 0.26  & 0.50  & 0.26  & 0.23  & 0.53  & 0.20  & 0.00  & 0.26 & 56.5 \\
      & ASVD & 0.12  & 0.26  & 0.51  & 0.26  & 0.22  & 0.52  & 0.19  & 0.00  & 0.26 & 56.5 \\
      & SVD-LLM & 0.16  & 0.33  & 0.54  & 0.29  & 0.23  & 0.56  & 0.21  & 0.00  & 0.29 & 63.0 \\
\cmidrule{2-12}          & Ours & 0.13  & 0.29  & 0.53  & 0.28  & 0.23  & 0.53  & 0.20  & 0.01  & 0.28 & 60.9 \\
      & Ours* & \textbf{0.18} & \textbf{0.4} & \textbf{0.56} & \textbf{0.35} & \textbf{0.26} & \textbf{0.61} & \textbf{0.21} & \textbf{0.02} & \textbf{0.32} & \textbf{69.6}\\
\bottomrule[2pt]
\end{tabular}%
}
\caption{Performance of LLaMA-7B compressed by GRASP (Ours* denotes the version with post-training compensation) and low-rank pruning baselines under 20\% to 50\% compression ratio on seven common sense reasoning datasets (measured by \textbf{accuracy\( \uparrow \)}) and GSM8K dataset (measured by \textbf{Exact Match Accuracy\( \uparrow \)}). Percentage represents the proportion of the original model's performance retained by the pruned method. The best performance is marked in bold.}
\label{tab:Comparison with SVD-based methods under different compression ratio}
\end{table*}

\subsection{Comparison with Pruning-based LLM Compression Methods}
\label{exp:Comparison with Pruning-based LLM Compression Methods}
\subsubsection{Comparison without Post-Training Compensation}
\paragraph{Models and Benchmarks.} In this experiment, we evaluate GRASP against representative structured pruning baselines using a more modern LLM architecture—LLaMA 3.1-8B-Instruct—\textbf{without applying any post-training compensation}. The model is compressed to 20\% of its original size and evaluated on seven commonsense reasoning benchmarks, including \textbf{WinoGrande}~\citep{Sakaguchi2020}, \textbf{HellaSwag}~\citep{zellers-etal-2019-hellaswag}, \textbf{OpenBookQA}~\citep{mihaylov-etal-2018-suit}, \textbf{PIQA}~\citep{bisk2020piqa}, \textbf{ARC-e}, \textbf{ARC-c}~\citep{clark2018think}, and \textbf{MathQA}~\citep{amini-etal-2019-mathqa}. All tasks are tested in a zero-shot
setting using the LM-Evaluation-Harness framework~(\citealp{eval-harness}).

\paragraph{Main Results.} As shown in Table~\ref{tab:Comparison with pruning-based methods without compensation},  GRASP achieves the highest average accuracy across seven commonsense reasoning benchmarks, consistently outperforming all baseline methods. In particular, GRASP improves over SliceGPT by 34\% in average accuracy and outperforms LaCo by 12\%. To further evaluate the generalizability of GRASP across different LLM architectures, we additionally conduct experiments on LLaMA 2-7B, LLaMA 2-13B, and Mistral-7B. Detailed results are presented in Appendix~\ref{appendix:Detailed Results on Common Sense Reasoning Benchmarks}. Notably, GRASP demonstrates significantly improved robustness across models, effectively mitigating the variability in pruning sensitivity across diverse architectures.

\subsubsection{Comparison with Post-Training Compensation}
\paragraph{Models and Benchmarks.} In this section, following prior research, we compress the LLaMA 2-7B model under a 25\% compression ratio and evaluate the compressed model on a broad set of natural language understanding (NLU) and question-answering (QA) benchmarks, including \textbf{CMNLI}~(\citealp{xu2020clue}), \textbf{HellaSwag}~(\citealp{zellers-etal-2019-hellaswag}), \textbf{PIQA}~(\citealp{bisk2020piqa}), \textbf{CHID}~(\citealp{zheng2019chid}), \textbf{WSC}~(\citealp{levesque2012winograd}), \textbf{CommonsenseQA}~(\citealp{talmor2018commonsenseqa}), \textbf{BoolQ}~(\citealp{clark2019boolq}), \textbf{MMLU}~(\citealp{hendrycks2020measuring}), \textbf{CMMLU}~(\citealp{li2023cmmlu}), \textbf{Race}~(\citealp{lai2017race}) and \textbf{C3}~(\citealp{sun2020investigating}). We utilized the OpenCompass evaluation framework~(\citealp{2023opencompass}) and report accuracy as the evaluation metric for all benchmarks under the PPL mode, following the same evaluation protocol as LaCo~\citep{yang2024laco}.

\paragraph{Main Results.} To ensure a fair comparison, we constrain the number of trainable parameters to remain approximately the same across all methods by retaining only 10\% of the parameters in each redundant layer and allowing only these parameters to be trainable. Notably, for GRASP's post-training compensation process, we fine-tune the compressed model on Alpaca~(\citealp{taori2023stanford}) for only one epoch to guarantee efficiency. Additional implementation details are provided in Appendix~\ref{appendix:Experimental Setup and Hyperparameters Configuration}. As shown in Table~\ref{tab:Comparison with pruning-based methods with compensation}, GRASP consistently outperforms the best-performing baselines LLM-Streamline on average and achieves a \textbf{94.9\%} of the original model performance at a 25\% compression ratio. In addition to achieving superior accuracy, GRASP also demonstrates more stable and faster convergence during post-training, which we attribute to its preservation of critical singular components.

\subsubsection{Comparison with Low-Rank Pruning Methods}
\paragraph{Models and Benchmarks.} In this section, we further compare our method against state-of-the-art structured low-rank pruning approaches—\textbf{FWSVD} (\citealp{hsu2022language}), \textbf{ASVD} (\citealp{yuan2023asvd}) and \textbf{SVD-LLM} (\citealp{wang2024svd}) on the LLaMA-7B model under various compression ratio. 8 datasets are used as evaluation benchmarks including seven commonsense reasoning datasets as Experiment 1 and one natural language generation~(NLG) benchmark \textbf{GSM8K}~(\citealp{cobbe2021training}). For all benchmarks, we report the zero-shot accuracy as the evaluation metric.

\paragraph{Main Results.} To evaluate the performance and stability of our proposed method, we conduct experiments under various compression ratios ranging from 20\% to 50\%. Table \ref{tab:Comparison with SVD-based methods under different compression ratio} summarizes the results for different methods. The results demonstrate that our proposed GRASP consistently outperforms the baseline methods on most benchmarks. Specifically, GRASP retains more than 91\% of the original performance at a 20\% compression ratio and 87\% under the compression ratio of 30\%. More importantly, with fast and resource-efficient post-training compensation, GRASP enables rapid accuracy recovery, achieving 70\% of the original model performance even at a 50\% compression ratio.

\begin{figure}[tp]
    \centering
    \includegraphics[width=1\linewidth]{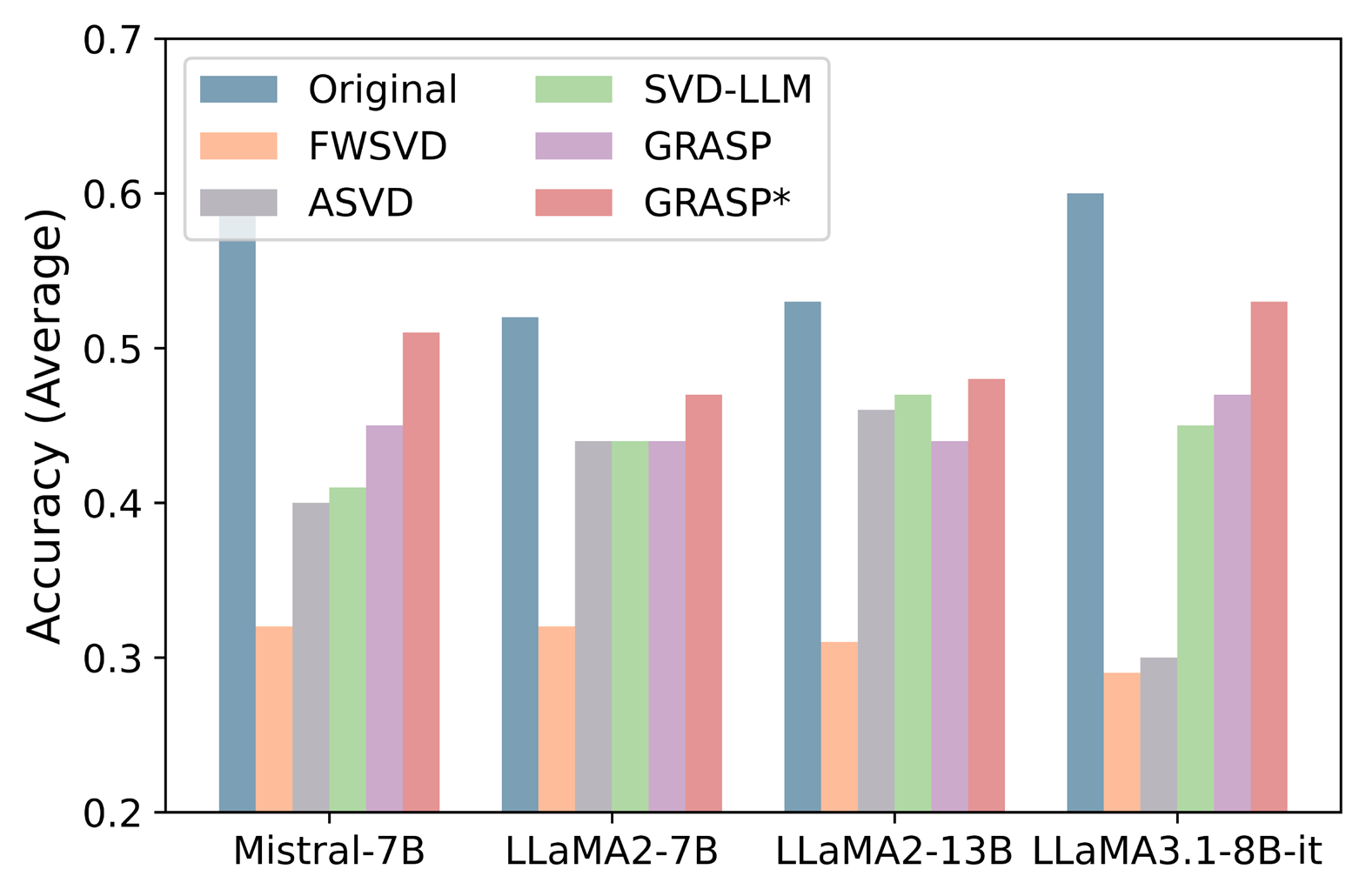}
    \caption{Comparison between our method and low-rank pruning baselines on four different LLMs. Average accuracy is reported across seven commonsense reasoning benchmarks: OpenBookQA, WinoGrande, HellaSwag, ARC-easy, ARC-challenge, PIQA, and MathQA.}
    \vspace{-5mm}
    \label{fig:accuracy}
\end{figure}

Furthermore, to assess the generalizability of GRASP across different LLM architectures, we compare its performance against structured low-rank pruning baselines under a 20\% compression ratio on four models from two distinct LLM families: LLaMA 2-7B, LLaMA 2-13B, LLaMA 3.1-8B-Instruct, and Mistral-7B. As shown in Figure~\ref{fig:accuracy}, GRASP consistently outperforms all baseline methods across architectures and exhibits greater robustness across different model families. The only exception is on LLaMA 2-13B, where GRASP slightly underperforms SVD-LLM. However, this performance gap can be quickly recovered through lightweight compensation. The detailed results are provided in Appendix~\ref{appedix:More Results on Other Models}.

\begin{figure*}[t]
    \centering
    \includegraphics[width=1\linewidth]{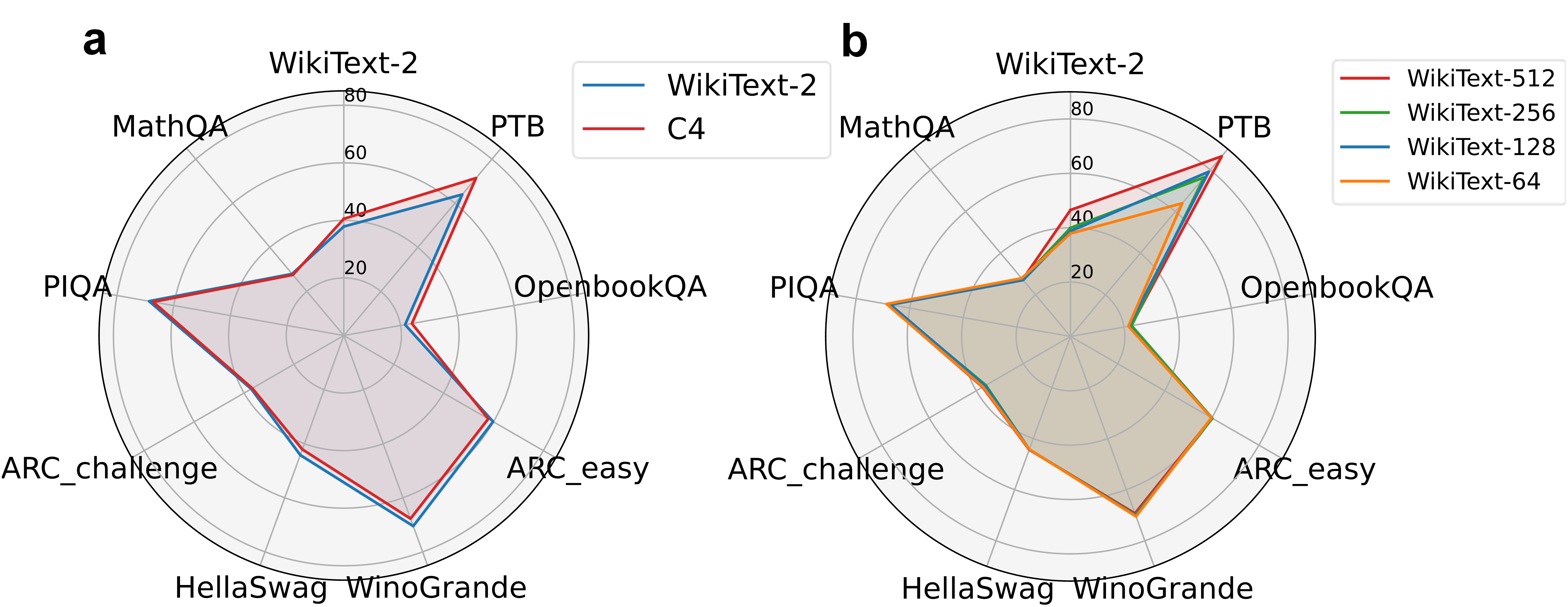}
    \caption{Performance of GRASP on LLaMA3.1-8B-Instruct under 20\% compression using (a) different calibration datasets (WikiText-2, C4) and (b) varying amounts of calibration data from WikiText-2. GRASP demonstrates limited sensitivity to calibration data changes, with final task performance varying within 4\%.}
    \label{fig:Ablation Study}
\end{figure*}

\subsection{Compression Costs and Inference Efficiency of GRASP}
\label{exp:Inference Efficiency of GRASP}
\begin{figure}[t]
    \centering
    \includegraphics[width=1\linewidth]{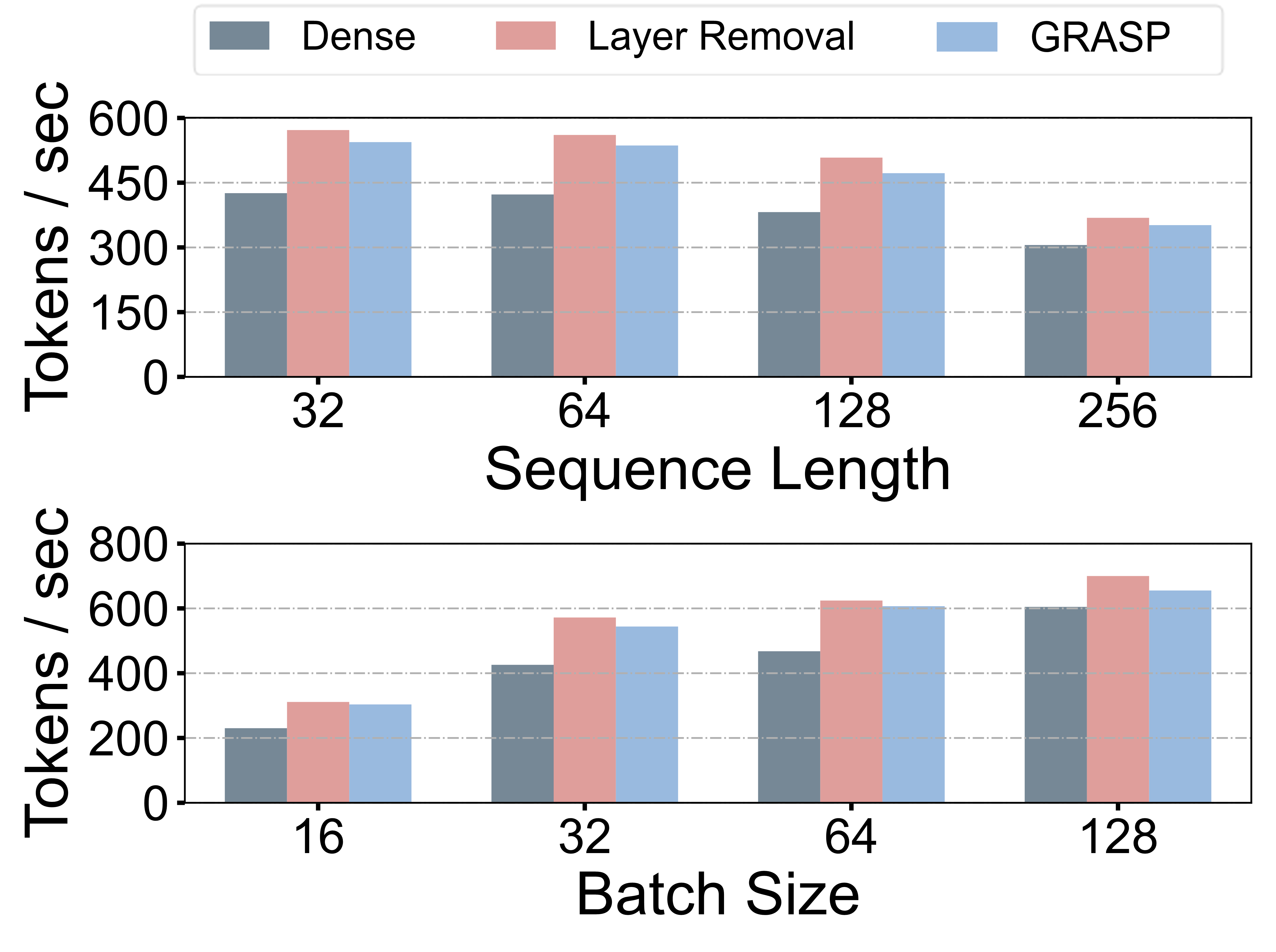}
    \caption{Throughput of LLaMA2-7B and GRASP compressed model under 25\% compression ratio on a single A100 GPU. \textbf{Top:} Throughput across different sequence lengths (batch size = 32). \textbf{Bottom:} Throughput across different batch sizes (sequence length = 32).}
    \label{fig:Inference}
\end{figure}

GRASP enables low-cost compression of LLMs while improving inference efficiency on real hardware. To evaluate its acceleration benefits, we measure the throughput (tokens per second) of the original LLaMA2-7B and its GRASP-compressed counterpart under varying sequence lengths and batch sizes. As shown in Figure~\ref{fig:Inference}, GRASP consistently improves generation speed and achieves acceleration comparable to direct layer removal. Notably, although GRASP retains a small subset of parameters within redundant layers to mitigate the performance drop caused by layer removal, these retained components are extremely low-rank and incur negligible inference overhead while preserving task performance. Additionally, we measure the compression time of GRASP and other structured pruning baselines when compressing LLaMA2-7B on an NVIDIA A100 GPU under a 25\% compression ratio. As reported in Table~\ref{tab:compression costs}, \textit{Pruning Time} refers to the time required to perform model compression up to the target sparsity, excluding any compensation. \textit{Compensation Time} refers to the time spent on post-compression procedures—such as fine-tuning or parameter updates—aimed at recovering model performance. The results show that GRASP is able to compress the model efficiently while maintaining strong performance, demonstrating its practicality for real-world deployment. 

\begin{table}[H]
\centering
\resizebox{0.95\linewidth}{!}{%
\begin{tabular}{c|c|c}
    \toprule[2pt]
    Model & Pruning Time (h) & Compensation Time (h) \\
    \midrule
    LaCO  & 0.05  & 1.2 \\
    SliceGPT & 0.6   & 0.76 \\
    LLM-Streamline & 0.03  & 0.7 \\
    GRASP & 0.16  & × \\
    GRASP* & 0.16  & 0.66 \\
    \bottomrule[2pt]
\end{tabular}%
}
\caption{Compression time of GRASP and structured pruning baselines on LLaMA2-7B under a 25\% compression ratio on a single A100 GPU. “×” indicates that GRASP does not require post-training compensation.}
\label{tab:compression costs}
\end{table}

\subsection{Ablation Study}
\label{sec:ablation_study}
\paragraph{Calibration Data.} We conduct ablation studies to assess the sensitivity of GRASP to the choice and size of the calibration dataset used for gradient-based singular value attribution. Table~\ref{tab:Ablation Study} and Figure~\ref{fig:Ablation Study} present the results on LLaMA3.1-8B-Instruct when varying the calibration dataset and the number of samples drawn from WikiText-2. The results indicate that GRASP remains robust across different calibration dataset choices and performs reliably even with limited calibration data. Remarkably, the method achieves strong performance with as few as 64 samples, demonstrating its effectiveness in low-data regimes.

\begin{table}[ht]
\centering
\resizebox{1\linewidth}{!}{%
\begin{tabular}{c|c|ccc}
\toprule[2pt]
Ablation Type & \begin{tabular}[c]{@{}c@{}}Calibration\\ Dataset\end{tabular} & WikiText-2 & PTB   & \begin{tabular}[c]{@{}c@{}}Average\\ Accuracy\end{tabular} \\
\hline
\multirow{2}[1]{*}{Varying Dataset} & WikiText-2 & 37.86 & 63.97 & 47.12 \\
      & C4    & 40.54 & 71.42 & 46.17 \\
\hline
\multirow{4}[2]{*}{Varying Number} & 64 & 46.5  & 86.51 & 47.06 \\
      & 128 & 39.91 & 76.41 & 46.93 \\
      & 256 & 38.73 & 79.13 & 46.67 \\
      & 512 & 37.86 & 63.97 & 47.12 \\
\bottomrule[2pt]
\end{tabular}%
}
\caption{Comparison of GRASP using different types and amounts of data for compressing LLaMA3.1-8B-Instruct at a 20\% compression ratio. Results are reported on the WikiText-2, PTB datasets~(measured by \textbf{perplexity\( \downarrow \)}) and the average accuracy across seven commonsense reasoning benchmarks: OpenBookQA, WinoGrande, HellaSwag, ARC-easy, ARC-challenge, PIQA, and MathQA.}
\vspace{-2mm}
\label{tab:Ablation Study}
\end{table}

\begin{figure}[ht]
    \centering
    \includegraphics[width=1\linewidth]{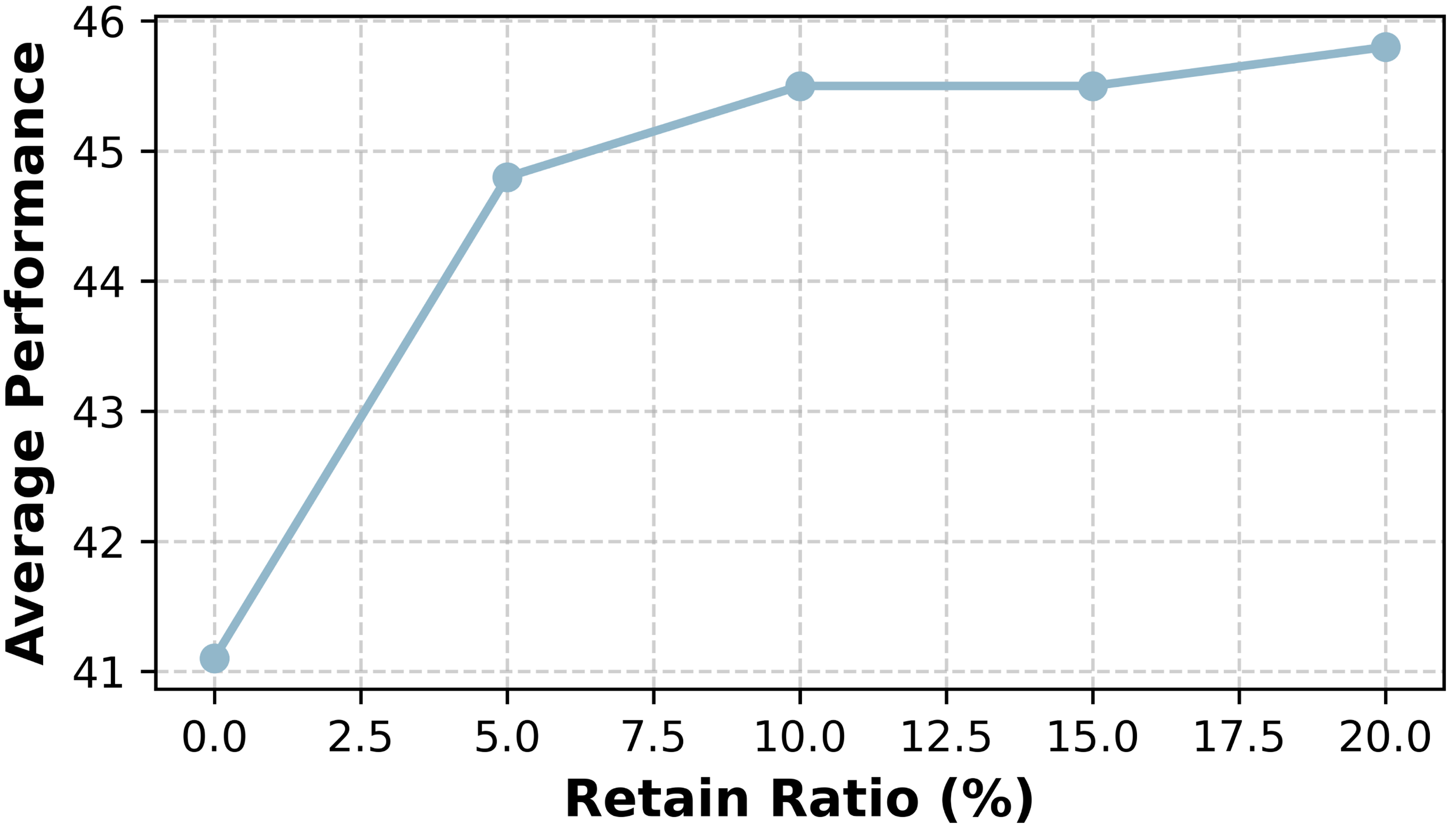}
    \caption{Effect of the retain ratio on average performance when compressing 8 redundant layers of LLaMA2-7B. The results show a sharp improvement up to a 10\% retain ratio, after which performance gains plateau, indicating 10\% as a practical inflection point.}
    \label{fig:trend}
    \vspace{-2mm}
\end{figure}

\paragraph{Retain Ratio \(r\)\%.}
\label{sec:ablation_retain_ratio}
We further explore the impact of the retain ratio \(r\) on GRASP’s performance. As shown in Table~\ref{tab:ablation_retain_ratio} and Figure~\ref{fig:trend}, performance improves rapidly when increasing \(r\) from 0\% to 10\%, after which the gains largely saturate. This indicates that a retain ratio of 10\% serves as a practical inflection point, providing a strong trade-off between compression and accuracy. In practice, we therefore adopt a fixed retain ratio of 10\%, which consistently yields stable performance across benchmarks while maintaining high compression efficiency. These results further confirm that the selected redundant layers contain substantial redundancy, as even a small fraction of retained singular components suffices to recover most of the original performance.

\begin{table*}[t]
\centering
\resizebox{0.9\linewidth}{!}{%
    \begin{tabular}{c|c|ccccccc}
    \toprule[2pt]
    Retain Ratio & \begin{tabular}[c]{@{}c@{}}Overall\\ Compression Ratio\end{tabular} & Openb. & ARC\_e & ARC\_c & WinoGrande & PIQA  & MathQA & Average \\
    \midrule
    0\%   & 24.0\% & 20.2  & 45.1  & 30.9  & 61.3  & 67.0  & 22.2  & 41.1 \\
    5\%   & 22.8\% & 22.4  & 54.9  & 31.8  & 66.0  & 70.0  & 23.8  & 44.8 \\
    10\%  & 21.6\% & 23.2  & 56.1  & 32.7  & 66.2  & 70.7  & 23.9  & 45.5 \\
    15\%  & 20.4\% & 23.4  & 56.3  & 32.9  & 65.6  & 70.7  & 24.1  & 45.5 \\
    20\%  & 19.2\% & 23.4  & 56.5  & 33.4  & 65.6  & 71.5  & 24.3  & 45.8 \\
    \bottomrule[2pt]
    \end{tabular}%
}
\caption{Ablation results on LLaMA2-7B by replacing 8 redundant layers with sensitivity-aware singular parameters while varying the retain ratio. The table reports the resulting overall compression ratio and downstream task performance across multiple benchmarks.}
\vspace{-3mm}
\label{tab:ablation_retain_ratio}
\end{table*}

\paragraph{One-shot Pruning vs Iterative Pruning} 
As detailed in Section \ref{method:Layer-wise Low-Rank Decomposition}, GRASP processes the redundant layers in a sequential manner, starting from the final redundant layer and proceeding backward through the network. This can be done using one-shot pruning, which identifies and decomposes all redundant layers in a single step, or iterative pruning, which processes layers one at a time to account for interactions between layers. We present the ablation results in Table \ref{tab:Pruning Strategy}, which shows that both approaches achieve similar performance, with one-shot pruning being more efficient.

\begin{table}[H]
\centering
\resizebox{\linewidth}{!}{%
\begin{tabular}{c|cccc}
\toprule[2pt]
Pruning Strategy & WikiText-2 & PTB & \begin{tabular}[c]{@{}c@{}}Average\\ Accuracy\end{tabular} & \begin{tabular}[c]{@{}c@{}}Compression\\ Time(h)\end{tabular} \\ \hline
One-shot Pruning & 37.86 & 63.97 & 47.12 & 0.16 \\
Iterative Pruning & 38.39 & 72.18 & 47.13 & 0.22 \\ 
\bottomrule[2pt]
\end{tabular}
}
\caption{Comparison of one-shot pruning and iterative pruning for LLaMA3.1-8B-Instruct under 20\% compression ratio.}
\vspace{-3mm}
\label{tab:Pruning Strategy}
\end{table}

\section{Discussion}
The effectiveness of GRASP stems from two key design principles that distinguish it from conventional structured pruning methods.

\paragraph{Leveraging Low-Rank Redundancy.} 
GRASP builds on the observation that the redundant layers in LLMs exhibit inherent low-rank characteristics. Rather than removing these layers entirely, which can lead to information loss and degraded performance, GRASP retains a small subset of critical singular directions identified through a gradient-based attribution mechanism. This approach allows the model to retain essential functional capacity while discarding non-informative parameters, leading to efficient yet accurate compression.

\paragraph{Preserving Informative Subspaces for Fast Recovery.} By maintaining only the most influential parameters within redundant layers, GRASP avoids the need for costly retraining while facilitating faster convergence when post-training compensation is applied. This contrasts with approaches like LLM-Streamline~(\citealp{chen2024compressing}) that insert randomly initialized or dense lightweight modules, which require more data and training time to approximate the original function. GRASP’s retention of critical low-rank components allows it to preserve the spectral alignment of the original model~(\citealp{oymak2019generalization}, \citealp{kamalakara2022exploring}, \citealp{shuttleworth2024lora}), resulting in more stable optimization dynamics and enhanced sample efficiency during post-training compensation.

\section{Related Work}
Structured pruning aims to reduce the size and computational cost of large language models (LLMs) by removing entire components such as layers, neurons, or dimensions, while maintaining model performance. Among this, \textbf{layer pruning} is a kind of structured pruning technique that eliminates redundant layers within Large Language Models (LLMs). Methods such as ShortGPT (\citealp{men2024shortgpt}) introduce a metric called Block Influence to assess the significance of individual layers, enabling efficient one-shot removal of less important layers. SLEB (\citealp{song2024sleb}) improves this by employing an iterative pruning strategy, evaluating the importance of each layer based on the current state of the layer-removed LLMs. LaCo (\citealp{yang2024laco}), on the other hand, proposes a gradual compression approach, progressively merging redundant layers from deeper to shallower parts of the network.

Although effective in reducing model size, layer pruning disrupts representation coherence, leading to performance degradation and increased perplexity, as analyzed by \cite{liang2024internal}. To mitigate this, \textbf{post-training compensation} methods have been proposed. \citealp{kim2024shortened} introduced Shortened LLaMA, which employs LoRA (\citealp{hu2022lora}) to restore pruned models' capabilities. However, LoRA modifies the singular value spectrum, potentially weakening pre-trained features (\citealp{shuttleworth2024lora}). LLM-Streamline (\citealp{chen2024compressing}) addresses this by training a lightweight module, such as an FFN or transformer layer, to approximate the removed layers. While effective, these methods impose high computational and data costs, limiting feasibility in resource-constrained settings.

Another line of structured pruning research focuses on low-rank approximation, where Singular Value Decomposition (SVD) is widely used to decompose weight matrices into low-rank structures, typically selecting top-k singular values based on Frobenius norm reconstruction loss. Recent methods have enhanced SVD to reduce LLM compression error. FWSVD \citep{hsu2022language} incorporates Fisher information to reweight the importance of parameters before applying SVD. ASVD \citep{yuan2023asvd} uses activation patterns from a calibration dataset to scale weight matrices, reducing compression-induced activation errors. SVD-LLM \citep{wang2024svd} applies truncation-aware data whitening and layer-wise updates to ensure a direct relationship between singular values and compression loss. Additionally, (\citealp{yu2023compressing}; \citealp{chavan-etal-2024-surgical}; \citealp{ji2024adaptive}) present another paradigm for low-rank compression of LLMs, where eigenvalue decomposition is applied to output activations, approximating the output activations with low-rank matrices. However, SVD-based methods require truncating at least 50\% of singular values to reduce parameters of square matrices (which are common in LLMs like Llama), which often leads to significant information loss. In contrast to these approaches, GRASP integrates gradient-based attribution into the low-rank decomposition process and replaces redundant layers with only a small fraction of parameters (typically 10\%), thereby enabling efficient compression while retaining critical information.

\section{Conclusion}
In this work, we proposed \textbf{GRASP}, a novel compression framework that replaces redundant layers in large language models with a small set of adaptively selected singular parameters. By leveraging the low-rank structure of redundant layers and incorporating gradient-based attribution, GRASP identifies critical components that preserve model functionality with minimal parameter overhead. It operates in a training-free manner and enables efficient post-training recovery with limited data. Extensive experiments across diverse LLM architectures and benchmarks demonstrate that GRASP consistently outperforms existing structured pruning methods in both accuracy and efficiency.

\section{Limitations}

While GRASP achieves competitive performance and compression efficiency, it also has several limitations that merit further exploration. 

First, our method relies on the assumption that layer redundancy can be identified via output similarity (e.g., cosine similarity between hidden states). While effective in practice, this heuristic may overlook more nuanced forms of redundancy that arise from distributed or task-specific behaviors within deeper model layers.

Second, GRASP depends on access to gradient information and a small calibration dataset to compute attribution scores. Although the data requirement is minimal and the method is training-free in its core form, this may limit applicability in strictly black-box or privacy-sensitive settings where gradients or internal representations are inaccessible. 

We also note that our experiments are conducted on models up to 13B parameters and primarily in English-language tasks. Extending GRASP to multilingual or much larger-scale models is a promising direction for future work, especially as the scale and diversity of LLMs continue to grow.

\section*{Ethical Considerations}
Our research adheres to the ACL Code of Ethics, ensuring transparency, responsible use of data, and consideration of potential social impacts. All datasets used in this work are publicly available and have been appropriately cited, ensuring compliance with data usage agreements and privacy regulations.

While GRASP is designed to optimize the efficiency and scalability of large language models, we recognize that such technologies could be misused in applications that may perpetuate harmful biases or deploy models in contexts lacking adequate oversight. To mitigate these risks, we advocate for responsible deployment practices, including thorough testing and monitoring for unintended biases.

Moreover, we acknowledge the computational resources required for training and testing large language models. To minimize environmental impact, we conducted experiments on energy-efficient hardware (NVIDIA A100 GPUs) and report our computational cost transparently. Further details can be found in the Appendix.

\bibliography{main}

@inproceedings{NEURIPS2020_1457c0d6,
 author = {Brown, Tom and Mann, Benjamin and Ryder, Nick and Subbiah, Melanie and Kaplan, Jared D and Dhariwal, Prafulla and Neelakantan, Arvind and Shyam, Pranav and Sastry, Girish and Askell, Amanda and Agarwal, Sandhini and Herbert-Voss, Ariel and Krueger, Gretchen and Henighan, Tom and Child, Rewon and Ramesh, Aditya and Ziegler, Daniel and Wu, Jeffrey and Winter, Clemens and Hesse, Chris and Chen, Mark and Sigler, Eric and Litwin, Mateusz and Gray, Scott and Chess, Benjamin and Clark, Jack and Berner, Christopher and McCandlish, Sam and Radford, Alec and Sutskever, Ilya and Amodei, Dario},
 booktitle = {Advances in Neural Information Processing Systems},
 editor = {H. Larochelle and M. Ranzato and R. Hadsell and M.F. Balcan and H. Lin},
 pages = {1877--1901},
 publisher = {Curran Associates, Inc.},
 title = {Language Models are Few-Shot Learners},
 volume = {33},
 year = {2020}
}

@article{zhou2024survey,
  title={A survey on efficient inference for large language models},
  author={Zhou, Zixuan and Ning, Xuefei and Hong, Ke and Fu, Tianyu and Xu, Jiaming and Li, Shiyao and Lou, Yuming and Wang, Luning and Yuan, Zhihang and Li, Xiuhong and others},
  journal={arXiv preprint arXiv:2404.14294},
  year={2024}
}

@article{frantar2022gptq,
  title={Gptq: Accurate post-training quantization for generative pre-trained transformers},
  author={Frantar, Elias and Ashkboos, Saleh and Hoefler, Torsten and Alistarh, Dan},
  journal={arXiv preprint arXiv:2210.17323},
  year={2022}
}

@article{lin2024awq,
  title={AWQ: Activation-aware Weight Quantization for On-Device LLM Compression and Acceleration},
  author={Lin, Ji and Tang, Jiaming and Tang, Haotian and Yang, Shang and Chen, Wei-Ming and Wang, Wei-Chen and Xiao, Guangxuan and Dang, Xingyu and Gan, Chuang and Han, Song},
  journal={Proceedings of Machine Learning and Systems},
  volume={6},
  pages={87--100},
  year={2024}
}

@inproceedings{xiao2023smoothquant,
  title={Smoothquant: Accurate and efficient post-training quantization for large language models},
  author={Xiao, Guangxuan and Lin, Ji and Seznec, Mickael and Wu, Hao and Demouth, Julien and Han, Song},
  booktitle={International Conference on Machine Learning},
  pages={38087--38099},
  year={2023},
  organization={PMLR}
}

@article{gu2023knowledge,
  title={Knowledge distillation of large language models},
  author={Gu, Yuxian and Dong, Li and Wei, Furu and Huang, Minlie},
  journal={arXiv preprint arXiv:2306.08543},
  year={2023}
}

@article{xu2024survey,
  title={A survey on knowledge distillation of large language models},
  author={Xu, Xiaohan and Li, Ming and Tao, Chongyang and Shen, Tao and Cheng, Reynold and Li, Jinyang and Xu, Can and Tao, Dacheng and Zhou, Tianyi},
  journal={arXiv preprint arXiv:2402.13116},
  year={2024}
}

@inproceedings{sun2024a,
  title={A Simple and Effective Pruning Approach for Large Language Models},
  author={Mingjie Sun and Zhuang Liu and Anna Bair and J Zico Kolter},
  booktitle={The Twelfth International Conference on Learning Representations},
  year={2024}
}

@article{ashkboos2024slicegpt,
  title={Slicegpt: Compress large language models by deleting rows and columns},
  author={Ashkboos, Saleh and Croci, Maximilian L and Nascimento, Marcelo Gennari do and Hoefler, Torsten and Hensman, James},
  journal={arXiv preprint arXiv:2401.15024},
  year={2024}
}

@article{men2024shortgpt,
  title={Shortgpt: Layers in large language models are more redundant than you expect},
  author={Men, Xin and Xu, Mingyu and Zhang, Qingyu and Wang, Bingning and Lin, Hongyu and Lu, Yaojie and Han, Xianpei and Chen, Weipeng},
  journal={arXiv preprint arXiv:2403.03853},
  year={2024}
}

@article{yang2024laco,
  title={Laco: Large language model pruning via layer collapse},
  author={Yang, Yifei and Cao, Zouying and Zhao, Hai},
  journal={arXiv preprint arXiv:2402.11187},
  year={2024}
}

@article{song2024sleb,
  title={SLEB: Streamlining LLMs through Redundancy Verification and Elimination of Transformer Blocks},
  author={Song, Jiwon and Oh, Kyungseok and Kim, Taesu and Kim, Hyungjun and Kim, Yulhwa and Kim, Jae-Joon},
  journal={arXiv preprint arXiv:2402.09025},
  year={2024}
}

@inproceedings{kim2024shortened,
  title={Shortened {LL}a{MA}: A Simple Depth Pruning for Large Language Models},
  author={Bo-Kyeong Kim and Geonmin Kim and Tae-Ho Kim and Thibault Castells and Shinkook Choi and Junho Shin and Hyoung-Kyu Song},
  booktitle={ICLR 2024 Workshop on Mathematical and Empirical Understanding of Foundation Models},
  year={2024}
}

@article{liang2024internal,
  title={Internal consistency and self-feedback in large language models: A survey},
  author={Liang, Xun and Song, Shichao and Zheng, Zifan and Wang, Hanyu and Yu, Qingchen and Li, Xunkai and Li, Rong-Hua and Wang, Yi and Wang, Zhonghao and Xiong, Feiyu and others},
  journal={arXiv preprint arXiv:2407.14507},
  year={2024}
}

@inproceedings{hu2022lora,
  title={Lo{RA}: Low-Rank Adaptation of Large Language Models},
  author={Edward J Hu and yelong shen and Phillip Wallis and Zeyuan Allen-Zhu and Yuanzhi Li and Shean Wang and Lu Wang and Weizhu Chen},
  booktitle={International Conference on Learning Representations},
  year={2022}
}

@article{chen2024compressing,
  title={Compressing large language models by streamlining the unimportant layer},
  author={Chen, Xiaodong and Hu, Yuxuan and Zhang, Jing},
  journal={arXiv preprint arXiv:2403.19135},
  year={2024}
}

@article{touvron2023llama1,
  title={Llama: Open and efficient foundation language models},
  author={Touvron, Hugo and Lavril, Thibaut and Izacard, Gautier and Martinet, Xavier and Lachaux, Marie-Anne and Lacroix, Timoth{\'e}e and Rozi{\`e}re, Baptiste and Goyal, Naman and Hambro, Eric and Azhar, Faisal and others},
  journal={arXiv preprint arXiv:2302.13971},
  year={2023}
}

@article{touvron2023llama,
  title={Llama 2: Open foundation and fine-tuned chat models},
  author={Touvron, Hugo and Martin, Louis and Stone, Kevin and Albert, Peter and Almahairi, Amjad and Babaei, Yasmine and Bashlykov, Nikolay and Batra, Soumya and Bhargava, Prajjwal and Bhosale, Shruti and others},
  journal={arXiv preprint arXiv:2307.09288},
  year={2023}
}

@article{dubey2024llama,
  title={The llama 3 herd of models},
  author={Dubey, Abhimanyu and Jauhri, Abhinav and Pandey, Abhinav and Kadian, Abhishek and Al-Dahle, Ahmad and Letman, Aiesha and Mathur, Akhil and Schelten, Alan and Yang, Amy and Fan, Angela and others},
  journal={arXiv preprint arXiv:2407.21783},
  year={2024}
}

@inproceedings{hsu2022language,
  title={Language model compression with weighted low-rank factorization},
  author={Yen-Chang Hsu and Ting Hua and Sungen Chang and Qian Lou and Yilin Shen and Hongxia Jin},
  booktitle={International Conference on Learning Representations},
  year={2022}
}

@article{yuan2023asvd,
  title={Asvd: Activation-aware singular value decomposition for compressing large language models},
  author={Yuan, Zhihang and Shang, Yuzhang and Song, Yue and Wu, Qiang and Yan, Yan and Sun, Guangyu},
  journal={arXiv preprint arXiv:2312.05821},
  year={2023}
}

@article{wang2024svd,
  title={Svd-llm: Truncation-aware singular value decomposition for large language model compression},
  author={Wang, Xin and Zheng, Yu and Wan, Zhongwei and Zhang, Mi},
  journal={arXiv preprint arXiv:2403.07378},
  year={2024}
}

@inproceedings{yu2023compressing,
  title={Compressing transformers: features are low-rank, but weights are not!},
  author={Yu, Hao and Wu, Jianxin},
  booktitle={Proceedings of the AAAI Conference on Artificial Intelligence},
  volume={37},
  pages={11007--11015},
  year={2023}
}

@inproceedings{chavan-etal-2024-surgical,
    title = "Surgical Feature-Space Decomposition of {LLM}s: Why, When and How?",
    author = "Chavan, Arnav  and
      Lele, Nahush  and
      Gupta, Deepak",
    editor = "Ku, Lun-Wei  and
      Martins, Andre  and
      Srikumar, Vivek",
    booktitle = "Proceedings of the 62nd Annual Meeting of the Association for Computational Linguistics (Volume 1: Long Papers)",
    month = aug,
    year = "2024",
    address = "Bangkok, Thailand",
    publisher = "Association for Computational Linguistics",
    pages = "2389--2400"
}

@inproceedings{ji2024adaptive,
  title={Adaptive Feature-based Low-Rank Compression of Large Language Models via Bayesian Optimization},
  author={Ji, Yixin and Xiang, Yang and Li, Juntao and Xia, Qingrong and Ye, Zi and Duan, Xinyu and Wang, Zhefeng and Chen, Kehai and Zhang, Min},
  booktitle={Findings of the Association for Computational Linguistics: EMNLP 2024},
  pages={4152--4168},
  year={2024}
}

@article{jiang2023mistral,
  title={Mistral 7B},
  author={Jiang, Albert Q and Sablayrolles, Alexandre and Mensch, Arthur and Bamford, Chris and Chaplot, Devendra Singh and Casas, Diego de las and Bressand, Florian and Lengyel, Gianna and Lample, Guillaume and Saulnier, Lucile and others},
  journal={arXiv preprint arXiv:2310.06825},
  year={2023}
}

@inproceedings{merity2017pointer,
  title={Pointer Sentinel Mixture Models},
  author={Stephen Merity and Caiming Xiong and James Bradbury and Richard Socher},
  booktitle={International Conference on Learning Representations},
  year={2017}
}

@article{2020t5,
  author  = {Colin Raffel and Noam Shazeer and Adam Roberts and Katherine Lee and Sharan Narang and Michael Matena and Yanqi Zhou and Wei Li and Peter J. Liu},
  title   = {Exploring the Limits of Transfer Learning with a Unified Text-to-Text Transformer},
  journal = {Journal of Machine Learning Research},
  year    = {2020},
  volume  = {21},
  number  = {140},
  pages   = {1-67}
}

@inproceedings{mihaylov-etal-2018-suit,
    title = "Can a Suit of Armor Conduct Electricity? A New Dataset for Open Book Question Answering",
    author = "Mihaylov, Todor  and
      Clark, Peter  and
      Khot, Tushar  and
      Sabharwal, Ashish",
    editor = "Riloff, Ellen  and
      Chiang, David  and
      Hockenmaier, Julia  and
      Tsujii, Jun{'}ichi",
    booktitle = "Proceedings of the 2018 Conference on Empirical Methods in Natural Language Processing",
    month = oct # "-" # nov,
    year = "2018",
    address = "Brussels, Belgium",
    publisher = "Association for Computational Linguistics",
    pages = "2381--2391"
}

@article{Sakaguchi2020, 
  title={WinoGrande: An Adversarial Winograd Schema Challenge at Scale},
  volume={34},
  number={05},
  journal={Proceedings of the AAAI Conference on Artificial Intelligence},
  author={Sakaguchi, Keisuke and Le Bras, Ronan and Bhagavatula, Chandra and Choi, Yejin},
  year={2020},
  month={Apr.},
  pages={8732-8740}
}

@inproceedings{zellers-etal-2019-hellaswag,
    title = "{H}ella{S}wag: Can a Machine Really Finish Your Sentence?",
    author = "Zellers, Rowan  and
      Holtzman, Ari  and
      Bisk, Yonatan  and
      Farhadi, Ali  and
      Choi, Yejin",
    editor = "Korhonen, Anna  and
      Traum, David  and
      M{\`a}rquez, Llu{\'\i}s",
    booktitle = "Proceedings of the 57th Annual Meeting of the Association for Computational Linguistics",
    month = jul,
    year = "2019",
    address = "Florence, Italy",
    publisher = "Association for Computational Linguistics",
    pages = "4791--4800"
}

@inproceedings{bisk2020piqa,
  title={Piqa: Reasoning about physical commonsense in natural language},
  author={Bisk, Yonatan and Zellers, Rowan and Gao, Jianfeng and Choi, Yejin and others},
  booktitle={Proceedings of the AAAI conference on artificial intelligence},
  volume={34},
  pages={7432--7439},
  year={2020}
}

@inproceedings{amini-etal-2019-mathqa,
    title = "{M}ath{QA}: Towards Interpretable Math Word Problem Solving with Operation-Based Formalisms",
    author = "Amini, Aida  and
      Gabriel, Saadia  and
      Lin, Shanchuan  and
      Koncel-Kedziorski, Rik  and
      Choi, Yejin  and
      Hajishirzi, Hannaneh",
    editor = "Burstein, Jill  and
      Doran, Christy  and
      Solorio, Thamar",
    booktitle = "Proceedings of the 2019 Conference of the North {A}merican Chapter of the Association for Computational Linguistics: Human Language Technologies, Volume 1 (Long and Short Papers)",
    month = jun,
    year = "2019",
    address = "Minneapolis, Minnesota",
    publisher = "Association for Computational Linguistics",
    pages = "2357--2367"
}

@article{clark2018think,
  title={Think you have solved question answering? try arc, the ai2 reasoning challenge},
  author={Clark, Peter and Cowhey, Isaac and Etzioni, Oren and Khot, Tushar and Sabharwal, Ashish and Schoenick, Carissa and Tafjord, Oyvind},
  journal={arXiv preprint arXiv:1803.05457},
  year={2018}
}

@misc{eval-harness,
  author={Gao, Leo and Tow, Jonathan and Abbasi, Baber and Biderman, Stella and Black, Sid and DiPofi, Anthony and Foster, Charles and Golding, Laurence and Hsu, Jeffrey and Le Noac'h, Alain and Li, Haonan and McDonell, Kyle and Muennighoff, Niklas and Ociepa, Chris and Phang, Jason and Reynolds, Laria and Schoelkopf, Hailey and Skowron, Aviya and Sutawika, Lintang and Tang, Eric and Thite, Anish and Wang, Ben and Wang, Kevin and Zou, Andy},
  title={A framework for few-shot language model evaluation},
  month=07,
  year=2024,
  publisher={Zenodo},
  version={v0.4.3},
}

@misc{taori2023stanford,
  title={Stanford alpaca: An instruction-following llama model},
  author={Taori, Rohan and Gulrajani, Ishaan and Zhang, Tianyi and Dubois, Yann and Li, Xuechen and Guestrin, Carlos and Liang, Percy and Hashimoto, Tatsunori B},
  year={2023}
}

@inproceedings{loshchilov2018decoupled,
  title={Decoupled Weight Decay Regularization},
  author={Ilya Loshchilov and Frank Hutter},
  booktitle={International Conference on Learning Representations},
  year={2019}
}

@article{shuttleworth2024lora,
  title={LoRA vs Full Fine-tuning: An Illusion of Equivalence},
  author={Shuttleworth, Reece and Andreas, Jacob and Torralba, Antonio and Sharma, Pratyusha},
  journal={arXiv preprint arXiv:2410.21228},
  year={2024}
}

@article{oymak2019generalization,
  title={Generalization guarantees for neural networks via harnessing the low-rank structure of the jacobian},
  author={Oymak, Samet and Fabian, Zalan and Li, Mingchen and Soltanolkotabi, Mahdi},
  journal={arXiv preprint arXiv:1906.05392},
  year={2019}
}

@article{kamalakara2022exploring,
  title={Exploring low rank training of deep neural networks},
  author={Kamalakara, Siddhartha Rao and Locatelli, Acyr and Venkitesh, Bharat and Ba, Jimmy and Gal, Yarin and Gomez, Aidan N},
  journal={arXiv preprint arXiv:2209.13569},
  year={2022}
}

@article{cobbe2021training,
  title={Training verifiers to solve math word problems},
  author={Cobbe, Karl and Kosaraju, Vineet and Bavarian, Mohammad and Chen, Mark and Jun, Heewoo and Kaiser, Lukasz and Plappert, Matthias and Tworek, Jerry and Hilton, Jacob and Nakano, Reiichiro and others},
  journal={arXiv preprint arXiv:2110.14168},
  year={2021}
}

@misc{hua2025dynamic,
  title={Dynamic low-rank estimation for transformer-based language models},
  author={Hua, Ting and Li, Xiao and Gao, Shangqian and Hsu, Yen-Chang and Shen, Yilin and Jin, Hongxia},
  year={2025},
  month=jan # "~23",
  publisher={Google Patents},
  note={US Patent App. 18/669,413}
}

@article{lecun1989optimal,
  title={Optimal brain damage},
  author={LeCun, Yann and Denker, John and Solla, Sara},
  journal={Advances in neural information processing systems},
  volume={2},
  year={1989}
}

@inproceedings{ma2023llmpruner,
    title={{LLM}-Pruner: On the Structural Pruning of Large Language Models},
    author={Xinyin Ma and Gongfan Fang and Xinchao Wang},
    booktitle={Thirty-seventh Conference on Neural Information Processing Systems},
    year={2023}
}

@article{xu2020clue,
  title={CLUE: A Chinese language understanding evaluation benchmark},
  author={Xu, Liang and Hu, Hai and Zhang, Xuanwei and Li, Lu and Cao, Chenjie and Li, Yudong and Xu, Yechen and Sun, Kai and Yu, Dian and Yu, Cong and others},
  journal={arXiv preprint arXiv:2004.05986},
  year={2020}
}

@article{zheng2019chid,
  title={ChID: A large-scale Chinese IDiom dataset for cloze test},
  author={Zheng, Chujie and Huang, Minlie and Sun, Aixin},
  journal={arXiv preprint arXiv:1906.01265},
  year={2019}
}

@article{levesque2012winograd,
  title={The Winograd schema challenge.},
  author={Levesque, Hector J and Davis, Ernest and Morgenstern, Leora},
  journal={KR},
  volume={2012},
  pages={13th},
  year={2012}
}

@article{talmor2018commonsenseqa,
  title={Commonsenseqa: A question answering challenge targeting commonsense knowledge},
  author={Talmor, Alon and Herzig, Jonathan and Lourie, Nicholas and Berant, Jonathan},
  journal={arXiv preprint arXiv:1811.00937},
  year={2018}
}

@article{clark2019boolq,
  title={Boolq: Exploring the surprising difficulty of natural yes/no questions},
  author={Clark, Christopher and Lee, Kenton and Chang, Ming-Wei and Kwiatkowski, Tom and Collins, Michael and Toutanova, Kristina},
  journal={arXiv preprint arXiv:1905.10044},
  year={2019}
}

@article{hendrycks2020measuring,
  title={Measuring massive multitask language understanding},
  author={Hendrycks, Dan and Burns, Collin and Basart, Steven and Zou, Andy and Mazeika, Mantas and Song, Dawn and Steinhardt, Jacob},
  journal={arXiv preprint arXiv:2009.03300},
  year={2020}
}

@article{li2023cmmlu,
  title={Cmmlu: Measuring massive multitask language understanding in chinese},
  author={Li, Haonan and Zhang, Yixuan and Koto, Fajri and Yang, Yifei and Zhao, Hai and Gong, Yeyun and Duan, Nan and Baldwin, Timothy},
  journal={arXiv preprint arXiv:2306.09212},
  year={2023}
}

@article{lai2017race,
  title={Race: Large-scale reading comprehension dataset from examinations},
  author={Lai, Guokun and Xie, Qizhe and Liu, Hanxiao and Yang, Yiming and Hovy, Eduard},
  journal={arXiv preprint arXiv:1704.04683},
  year={2017}
}

@article{sun2020investigating,
  title={Investigating prior knowledge for challenging chinese machine reading comprehension},
  author={Sun, Kai and Yu, Dian and Yu, Dong and Cardie, Claire},
  journal={Transactions of the Association for Computational Linguistics},
  volume={8},
  pages={141--155},
  year={2020},
  publisher={MIT Press One Rogers Street, Cambridge, MA 02142-1209, USA journals-info~…}
}

@misc{2023opencompass,
    title={OpenCompass: A Universal Evaluation Platform for Foundation Models},
    author={OpenCompass Contributors},
    howpublished = {\url{https://github.com/open-compass/opencompass}},
    year={2023}
}

@inproceedings{frantar2023sparsegpt,
  title={Sparsegpt: Massive language models can be accurately pruned in one-shot},
  author={Frantar, Elias and Alistarh, Dan},
  booktitle={International conference on machine learning},
  pages={10323--10337},
  year={2023},
  organization={PMLR}
}

@article{sun2023simple,
  title={A simple and effective pruning approach for large language models},
  author={Sun, Mingjie and Liu, Zhuang and Bair, Anna and Kolter, J Zico},
  journal={arXiv preprint arXiv:2306.11695},
  year={2023}
}

\clearpage
\appendix
\section{Appendix}
\subsection{The gradient of singular values}
\label{appendix:Proof}

For a weight matrix \(W \in \mathbb{R}^{m \times n}\) in the selected redundant layers, its differential form can be expressed as:
\begin{equation*}
\partial W = \partial U \Sigma V^T + U \partial \Sigma V^T + U \Sigma \partial V^T
\end{equation*}
\begin{equation*}
U^T \partial W V = U^T \partial U \Sigma + \partial \Sigma + \Sigma V^T \partial V
\end{equation*}
Since both \(U\) and \(V\) are orthogonal matrices, we have:
\begin{equation*}
U^T U = I_m, \quad V^T V = I_n
\end{equation*}
\begin{equation*}
\partial U^T U + U^T \partial U = O_m, \quad \partial V^T V + V^T \partial V = O_n
\end{equation*}
This implies that \(U^T dU\) and \(dV^T V\) are asymmetric matrices. Therefore, the diagonal elements of \(U^T dU \Sigma\) and \(\Sigma V^T dV\) are zero, leading to the diagonal elements of \(U^T \partial W V\) being:
\begin{equation*}
I_k \odot U^T \partial W V = \partial \Sigma
\end{equation*}
where $I_k$ represents the \(k \times k\) identity matrix, \(\odot\) denotes element-wise multiplication.

For a singular value \(\sigma_i\), its differential form can be written as:
\begin{equation*}
    \partial{\sigma_i} = u_i^T \partial W v_i 
\end{equation*}
Since \(\sigma_i\) is a scalar, we have:
\begin{align*}
    \partial \sigma_i &= \text{tr}(\partial \sigma_i) \\
    &= \text{tr}(u_i^T \partial W v_i) \\
    &= \text{tr}[(u_i v_i^T)^T \partial W]
\end{align*}
thereby, the derivative of \(\sigma_i\) with respect to W is:
\begin{equation*}
    \frac{\partial \sigma_i}{\partial W} = u_iv_i^T
\end{equation*}

For a calibration dataset \(D\), the gradient of a singular value \(\sigma_i\) with respect to the task loss can be interpreted as the projection of the weight gradient matrix \(G\) onto the corresponding singular direction, given by:
\begin{equation*}
    \frac{\partial L}{\partial \sigma_i} = u_i^T \frac{\partial L}{\partial W} v_i
\end{equation*}
Then, for all the singular values \(\Sigma\), we have: 
\begin{equation*}
\frac{\partial L}{\partial \Sigma} = I_k \odot U^T \left( \frac{\partial L}{\partial W} \right) V
\end{equation*}

\subsection{Experimental Setup and Hyperparameters Configuration}
\label{appendix:Experimental Setup and Hyperparameters Configuration}
To ensure a fair comparison, all experimental setups are consistent across all methods. In the following, we describe the experimental setup and hyperparameters configuration in detail.

\paragraph{Hyperparameters Configurations} For post-training compensation, all models compressed by GRASP are trained on the Alpaca (\citealp{taori2023stanford}) dataset for 1 epoch with a batch size of 32. We use AdamW (\citealp{loshchilov2018decoupled}) as our optimizer and set the learning rate to $3 \times 10^{-4}$. All our experiments are conducted on a single A100 GPU with mixed precision enabled. Table \ref{tab:Experimental Setup and Hyperparameter Configuration} provides the detailed configurations of post-training compensation. 

\begin{table}[H]
\centering
\resizebox{1\linewidth}{!}{%
\begin{tabular}{c|cl}
\cline{1-2}
HyperParameters & Setting &  \\ \cline{1-2}
Dataset & Alpaca &  \\
Huggingface Dataset Path & yahma/alpaca-cleaned &  \\
Batch Size & 32 &  \\
Micro Batch Size & 4 &  \\
Epochs & 1 &  \\
Learning Rate & 3.00E-04 &  \\
Max Length & 256 &  \\
Train on Inputs & TRUE &  \\
Add EOS Token & FALSE &  \\
LoRA-Rank & 256 &  \\
LoRA-Alpha & 16 &  \\
LoRA-Dropout & 0.05 &  \\
LoRA-Target-Modules & \begin{tabular}[c]{@{}c@{}}q\_proj, k\_proj, v\_proj, o\_proj,\\ up\_proj, down\_proj, gate\_proj\end{tabular} &  \\
Prompt-Template & Alpaca Template &  \\ \cline{1-2}
\end{tabular}
}
\caption{Experimental setup and hyperparameters configurations.}
\label{tab:Experimental Setup and Hyperparameter Configuration}
\end{table}

\begin{table*}[t]
\centering
\resizebox{\linewidth}{!}{%
\begin{tabular}{c|c|c|c|c|c|c}
\toprule[2pt]
\textbf{Method} & \textbf{Metric} & \textbf{Calibration Data} & \textbf{Need Post-Training} & \textbf{Training Data} & \textbf{Training Dataset Size} & \textbf{Training Module} \\
\midrule
ShortGPT & Cosine Similarity & WikiText-2 & No & None  & None  & None \\
\midrule
LaCo  & Cosine Similarity & WikiText-2 & Optional & Unpublished & 1B    & Full Parameters \\
\midrule
SliceGPT & PCA & \begin{tabular}[c]{@{}c@{}}WikiText-2\\ Alpaca\end{tabular} & Optional & Alpaca & 5k  & Full Parameters \\
\midrule
Shortened LLaMA & \begin{tabular}[c]{@{}c@{}}Taylor\\ Perplexity\end{tabular} & BookCorpus & Optional & \begin{tabular}[c]{@{}c@{}}SlimPajama\\ Alpaca\end{tabular} & \begin{tabular}[c]{@{}c@{}}627B\\ 50k\end{tabular} & \begin{tabular}[c]{@{}c@{}}Full Parameters\\ LoRA-Adapter\end{tabular} \\
\midrule
LLM-Streamline & Cosine Similarity & WikiText-2 & Yes   & SlimPajama & 30k   & \begin{tabular}[c]{@{}c@{}}Lightweight\\ Network\end{tabular} \\
\midrule
GRASP & Cosine Similarity & WikiText-2 & Optional & Alpaca & 50k   & \begin{tabular}[c]{@{}c@{}}Low-rank\\ Modules\end{tabular} \\
\bottomrule[2pt]
\end{tabular}%
}
\caption{Comparison of pruning-based LLM compression methods, where the metric indicates the criterion used to identify redundant modules. "Optional" refers to methods that can either work without post-training or recover performance through post-training. Shortened LLaMA consists of two training stages: initial continual pre-training on the SlimPajama dataset, followed by LoRA fine-tuning on the Alpaca dataset.}

\label{tab:Comparison of different pruning-based LLM Compression Methods}
\end{table*}

\subsection{Relationship between Singular Group Retail Ratio and Target Overall Compression Ratio}
\label{appendix:retain_ratio}
In GRASP, the retain ratio \(r\)\% denotes the proportion of singular components preserved within each weight matrix of the selected redundant layers. The target overall model compression ratio, which is jointly determined by (i) the number of layers selected in Step 1, and (ii) the retain ratio \(r\)\% applied in Step 2.

In practice, we adopt a fixed retain ratio—typically 10\%—as ablation studies~(Section~\ref{sec:ablation_retain_ratio}) show it achieves a favorable trade-off between compression and performance. Table~\ref{tab:ablation_retain_ratio}  further illustrates how varying the retain ratio influences the overall compression ratio and downstream performance across several benchmarks when compressing 8 layers of LLaMA2-7B. These results highlight the high redundancy of the selected layers: even a small fraction of retained singular components is sufficient to recover most of the original performance.

\subsection{Comparison of Concurrent Structured Pruning Methods}
\label{appendix:Comparison of Concurrent Pruning-based Methods}
We provide a detailed comparison of concurrent structured pruning LLM compression methods, and the results are summarized in Table \ref{tab:Comparison of different pruning-based LLM Compression Methods}.

\subsection{Detailed Results on Commonsense Reasoning Benchmarks}
\label{appendix:Detailed Results on Common Sense Reasoning Benchmarks}
In this section, we provide detailed results for GRASP and baseline methods on the commonsense reasoning benchmarks using LLaMA2-7B, LLaMA2-13B, and Mistral-7B. These results extend the main experiments presented in Table~\ref{tab:Comparison with pruning-based methods without compensation}, offering a comprehensive view of GRASP's performance across different model architectures. 

Table~\ref{tab:pruning-based without post-training Performance} reports accuracy without post-training compensation. The results demonstrate that GRASP consistently maintains strong accuracy and demonstrates robustness across model scales and families.

\subsection{More Results on Other Models}
\label{appedix:More Results on Other Models}
To further validate the generalizability of GRASP across diverse LLM architectures, we present additional results under a 20\% compression ratio on three representative models beyond LLaMA-7B, LLaMA2-13B, LLaMA3.1-8B-Instruct, and Mistral-7B. Table~\ref{tab:Comparison with SVD-based methods on Different LLMs} summarizes the performance comparison between GRASP and structured low-rank pruning baselines, including FWSVD, ASVD, and SVD-LLM, across eight evaluation benchmarks.

Consistent with our findings in the main text, GRASP achieves superior or comparable accuracy across most benchmarks. While SVD-LLM slightly outperforms GRASP on LLaMA2-13B, our method demonstrates stronger robustness across model families and benefits from more stable post-compensation recovery. These results highlight GRASP’s effectiveness as a generalizable and architecture-agnostic compression strategy.


\begin{table*}[t]
\centering
\resizebox{0.98\linewidth}{!}{%
\begin{tabular}{c|c|c|ccccccc|c}
\toprule[2pt]
Model & Method & Ratio  & Openb. & ARC\_e & WinoG. & HeSW & ARC\_c & PIQA  & MathQA & Average \\
\midrule
\multirow{5}[2]{*}{Mistral-7B-v0.1} & \cellcolor[rgb]{ .91,  .91,  .91}Dense & \cellcolor[rgb]{ .91,  .91,  .91}0.0\% & \cellcolor[rgb]{ .91,  .91,  .91}0.33 & \cellcolor[rgb]{ .91,  .91,  .91}0.81 & \cellcolor[rgb]{ .91,  .91,  .91}0.74 & \cellcolor[rgb]{ .91,  .91,  .91}0.61 & \cellcolor[rgb]{ .91,  .91,  .91}0.50 & \cellcolor[rgb]{ .91,  .91,  .91}0.81 & \cellcolor[rgb]{ .91,  .91,  .91}0.36 & \cellcolor[rgb]{ .91,  .91,  .91}0.59 \\
      & LaCo  & 21.1\% & 0.20  & 0.35  & 0.58  & 0.26  & 0.25  & 0.53  & 0.24  & 0.34 \\
      & ShortGPT & 21.1\% & 0.19  & \textbf{0.57} & \textbf{0.68} & \textbf{0.46} & 0.37  & \textbf{0.71} & \textbf{0.26} & \textbf{0.46} \\
      & SliceGPT & 20.0\% & 0.19  & 0.51  & 0.59  & 0.35  & 0.25  & 0.61  & 0.23  & 0.39 \\
      & GRASP & 20.0\% & \textbf{0.21} & 0.56  & \textbf{0.68} & 0.43  & \textbf{0.38} & 0.67  & \textbf{0.26} & \textbf{0.46} \\
\midrule
\multirow{5}[2]{*}{LLaMA2-7B} & \cellcolor[rgb]{ .91,  .91,  .91}Dense & \cellcolor[rgb]{ .91,  .91,  .91}0.0\% & \cellcolor[rgb]{ .91,  .91,  .91}0.32 & \cellcolor[rgb]{ .91,  .91,  .91}0.69 & \cellcolor[rgb]{ .91,  .91,  .91}0.67 & \cellcolor[rgb]{ .91,  .91,  .91}0.57 & \cellcolor[rgb]{ .91,  .91,  .91}0.40 & \cellcolor[rgb]{ .91,  .91,  .91}0.78 & \cellcolor[rgb]{ .91,  .91,  .91}0.28 & \cellcolor[rgb]{ .91,  .91,  .91}0.53 \\
      & LaCo  & 18.1\% & \textbf{0.26} & 0.48  & 0.59  & 0.42  & 0.32  & 0.69  & \textbf{0.24} & 0.43 \\
      & ShortGPT & 21.1\% & 0.23  & 0.49  & \textbf{0.63} & 0.42  & 0.31  & 0.68  & 0.23  & 0.43 \\
      & SliceGPT & 21.5\% & 0.22  & \textbf{0.54} & 0.61  & 0.37  & 0.28  & 0.63  & 0.23  & 0.41 \\
      & GRASP & 21.6\% & 0.24  & \textbf{0.54} & \textbf{0.63} & \textbf{0.43} & \textbf{0.33} & \textbf{0.71} & 0.23  & \textbf{0.44} \\
\midrule
\multirow{5}[1]{*}{LLaMA-2-13B} & \cellcolor[rgb]{ .91,  .91,  .91}Dense & \cellcolor[rgb]{ .91,  .91,  .91}0.0\% & \cellcolor[rgb]{ .91,  .91,  .91}0.32 & \cellcolor[rgb]{ .91,  .91,  .91}0.73 & \cellcolor[rgb]{ .91,  .91,  .91}0.70 & \cellcolor[rgb]{ .91,  .91,  .91}0.60 & \cellcolor[rgb]{ .91,  .91,  .91}0.46 & \cellcolor[rgb]{ .91,  .91,  .91}0.79 & \cellcolor[rgb]{ .91,  .91,  .91}0.30 & \cellcolor[rgb]{ .91,  .91,  .91}0.56 \\
      & LaCo  & 19.5\% & 0.28  & 0.52  & 0.63  & 0.43  & 0.33  & 0.70  & \textbf{0.25} & 0.45 \\
      & ShortGPT & 22.1\% & 0.24  & 0.50  & 0.63  & 0.46  & 0.33  & 0.7   & 0.24  & 0.44 \\
      & SliceGPT & 20.0\% & \textbf{0.29} & 0.59  & 0.65  & 0.39  & 0.32  & 0.64  & 0.24  & 0.45 \\
      & GRASP & 20.0\% & 0.26  & \textbf{0.61} & \textbf{0.66} & \textbf{0.47} & \textbf{0.35} & \textbf{0.73} & 0.24  & \textbf{0.47} \\
\midrule
\multirow{5}[1]{*}{LLaMA3.1-8B-Instruct} & \cellcolor[rgb]{ .91,  .91,  .91}Dense & \cellcolor[rgb]{ .91,  .91,  .91}0.0\% & \cellcolor[rgb]{ .91,  .91,  .91}0.34 & \cellcolor[rgb]{ .91,  .91,  .91}0.82 & \cellcolor[rgb]{ .91,  .91,  .91}0.74 & \cellcolor[rgb]{ .91,  .91,  .91}0.59 & \cellcolor[rgb]{ .91,  .91,  .91}0.52 & \cellcolor[rgb]{ .91,  .91,  .91}0.80 & \cellcolor[rgb]{ .91,  .91,  .91}0.39 & \cellcolor[rgb]{ .91,  .91,  .91}0.60 \\
      & LaCo  & 19.0\% & \textbf{0.26} & 0.49  & 0.65  & 0.33  & 0.30  & 0.65  & \textbf{0.30} & 0.42 \\
      & ShortGPT & 21.7\% & 0.21  & 0.57  & 0.66  & 0.42  & 0.32  & 0.67  & 0.26  & 0.44 \\
      & SliceGPT & 20.0\% & 0.15  & 0.43  & 0.51  & 0.30  & 0.23  & 0.58  & 0.22  & 0.35 \\
      & GRASP & 20.0\% & 0.22  & \textbf{0.60} & \textbf{0.70} & \textbf{0.44} & \textbf{0.37} & \textbf{0.69} & 0.28  & \textbf{0.47} \\
\bottomrule[2pt]
\end{tabular}%
}
\caption{Zero-shot performance of GRASP and pruning-based without post-training baselines under 20\% compression ratio. Results are reported on seven reasoning datasets (individual and average accuracy). Bold values indicate the best performance.}
\label{tab:pruning-based without post-training Performance}
\end{table*}

\begin{table*}[t]
\centering
\resizebox{0.96\linewidth}{!}{%
\begin{tabular}{c|cc|cc|cc|cc}
\toprule[2pt]
      & \multicolumn{2}{c|}{\textbf{Mistral-7B}} & \multicolumn{2}{c|}{\textbf{LLaMA2-7B}} & \multicolumn{2}{c|}{\textbf{LLaMA2-13B}} & \multicolumn{2}{c}{\textbf{LLaMA3.1-8B-Instruct}} \\
\hline
Method & \begin{tabular}[c]{@{}c@{}}PPL \( \downarrow \)\\ (WikiText-2)\end{tabular} & \begin{tabular}[c]{@{}c@{}}Acc \( \uparrow \)\\ Average\end{tabular} & \begin{tabular}[c]{@{}c@{}}PPL \( \downarrow \)\\ (WikiText-2)\end{tabular} & \begin{tabular}[c]{@{}c@{}}Acc \( \uparrow \)\\ Average\end{tabular} & \begin{tabular}[c]{@{}c@{}}PPL \( \downarrow \)\\ (WikiText-2)\end{tabular} & \begin{tabular}[c]{@{}c@{}}Acc \( \uparrow \)\\ Average\end{tabular} & \begin{tabular}[c]{@{}c@{}}PPL \( \downarrow \)\\ (WikiText-2)\end{tabular} & \begin{tabular}[c]{@{}c@{}}Acc \( \uparrow \)\\ Average\end{tabular} \\
\hline
Original & 5.25  & 0.59  & 5.68  & 0.52  & 5.47  & 0.53  & 7.21  & 0.60 \\
\hline
FWSVD & 6357  & 0.32  & 1727  & 0.32  & 2360  & 0.31  & 3256.7 & 0.29 \\
ASVD  & 19.28 & 0.4   & 11.14 & 0.44  & 9.70  & 0.46  & 2443.99 & 0.30 \\
SVD-LLM\textsuperscript{\textdagger} & \textbf{10.21} & 0.41  & \textbf{7.94} & 0.44  & \textbf{8.50} & 0.47  & -     & - \\
\hline
Ours  & 18.42 & 0.45  & 14.79 & 0.44  & 16.12 & 0.44  & 37.86 & 0.47 \\
Ours* & 11.62 & \textbf{0.51} & 10.19 & \textbf{0.47} & 9.59  & \textbf{0.48} & \textbf{14.13} & \textbf{0.53} \\
\bottomrule[2pt]
\end{tabular}%
}
\caption{Perplexity(\( \downarrow \)) of GRASP and low-rank pruning baselines on the WikiText-2 datasets and the average accuracy(\( \uparrow \)) on seven common sense reasoning datasets for four different LLMs under 20\% compression ratio. "\textdagger" indicates that we refer to the results in the original paper. The best performance is marked in bold.}
\vspace{-5mm}
\label{tab:Comparison with SVD-based methods on Different LLMs}.
\end{table*}

\subsection{Evaluation Results on LongBench}
\label{appendix:Evaluation Results on LongBench}
\vspace{-1mm}
In this section, we present the detailed results of LLaMA3.1-8B-Instruct and its compressed version by GRASP under 20\% compression ratio on LongBench, which are presented in Table \ref{tab:Performance on LongBench}. The results illustrate that GRASP with post-training compensation still maintains superior performance on long-form reasoning and complex generative tasks.

\subsection{Robustness of GRASP towards different calibration Dataset}
\label{appendix:Robustness of GRASP towards different representation evaluation dataset}

In this section, we provide details of the ablation studies conducted to investigate the impact of calibration datasets and the amount of data used for singular value gradient attribution. Specifically, we selected 512 samples from WikiText-2 (\citealp{merity2017pointer}) and C4 (\citealp{2020t5}) as calibration data to assess the performance of GRASP when compressing LLaMA3.1-8B-Instruct under 20\% compression ratio. Additionally, we selected 64, 128, 256 and 512 samples from WikiText-2 to examine the robustness of GRASP to the change in the number of calibration data. All calibration data were randomly selected from the training splits of the downstream datasets, ensuring no data leakage. As shown in Figure \ref{fig:Ablation Study},  we can observe that GRASP consistently achieves strong performance, indicating that our method is robust to variations in both the calibration dataset and the number of data. Tables \ref{tab:Robustness of GRASP towards different representation evaluation dataset} and \ref{tab:Robustness of GRASP towards different number of calibration data} summarize the results of GRASP when compressing LLaMA3.1-8B-Instruct with different calibration datasets (WikiText-2, C4) and varying numbers of calibration data.

\subsection{Comparison with Unstructured Pruning Methods}
To provide a more comprehensive picture of model compression, we have included comparisons with two widely used unstructured pruning methods, SparseGPT~(\citealp{frantar2023sparsegpt}) and Wanda~(\citealp{sun2023simple}) in our experiments. While unstructured pruning introduces fine-grained sparsity in weight matrices, it typically requires dedicated sparse formats to realize storage or runtime benefits, which complicates direct comparison under the same compression ratio. To ensure a fair evaluation, we adopt the commonly used 2:4 sparsity pattern (i.e., 50\% sparsity) for both methods, and compare them against GRASP at a 20\% structured compression ratio. The results are summarized in Table~\ref{tab:Comparison_with_unstructured_pruning_methods}, indicating that GRASP achieves competitive or superior performance against unstructured pruning methods across all benchmarks.

\begin{table*}[t]
\centering
\resizebox{0.95\linewidth}{!}{%
\begin{tabular}{c|cccc|cccc|cc|c}
\toprule[2pt]
\multirow{2}[4]{*}{Model} & \multicolumn{4}{c|}{Summarization} & \multicolumn{4}{c|}{Few-shot Learning} & \multicolumn{3}{c}{Synthetic Task} \\
\cmidrule{2-12}          & \textbf{1-1} & \textbf{1-2} & \textbf{1-3} & \textbf{1-4} & \textbf{1-1}  & \textbf{2-1} & \textbf{2-2} & \textbf{2-3}  & \textbf{2-4} & \multicolumn{1}{c}{\textbf{3-1}} & \textbf{3-2} \\
\midrule
LLaMA3.1-8B-Instruct & 28.35 & 20.04 & 25.85 & 14.51 & 63    & 51.3  & 39.29 & 16.5  & 1.64  & \multicolumn{1}{c}{9.67} & 17.17 \\
GRASP & \textbf{25.76} & \textbf{19.41} & 25.97 & \textbf{8.99} & \textbf{59.5} & 67.44 & \textbf{38.41} & \textbf{18} & 1     & \multicolumn{1}{c}{\textbf{10}} & \textbf{18} \\
\midrule[2pt]
\midrule[2pt]
\multirow{2}[4]{*}{Model} & \multicolumn{4}{c|}{One-Doc QA} & \multicolumn{4}{c|}{Multi-Doc QA} & \multicolumn{2}{c|}{Code Completion} & Average \\
\cmidrule{2-12}          & \textbf{4-1} & \textbf{4-2} & \textbf{4-3} & \textbf{4-4} & \textbf{5-1} & \textbf{5-2} & \textbf{5-3} & \textbf{5-4} & \textbf{6-1}  & \textbf{6-2} & \textbf{ALL} \\
\midrule
LLaMA3.1-8B-Instruct & 9.96  & 6.04  & 18.75 & 14.27 & 11.27 & 11.57 & 7.65  & 34.16 & 56.47 & 51.34 & 26.09 \\
GRASP & 15.32 & \textbf{20.47} & 34.52 & 16.46 & 31.33 & \textbf{26.18} & \textbf{10.28} & \textbf{29.86} & \textbf{33.74} & \textbf{38.09} & 26.26 \\
\bottomrule[2pt]
\end{tabular}%
}
\caption{Performance comparison of LLaMA3.1-8B-Instruct and its compressed version by GRASP under 20\% compression ratio on LongBench. The datasets are grouped as follows: 
\textbf{(1-1 to 1-4)} denote GovReport, QMSum, MultiNews, and VCSUM;
\textbf{(2-1 to 2-4)} denote TREC, TriviaQA, SAMSum, and LSHT;
\textbf{(3-1 to 3-3)} denote PassageCount, PassageRetrieval-en, and PassageRetrieval-zh;
\textbf{(4-1 to 4-4)} denote NarrativeQA, Qasper, MultiFieldQA-en, and MultiFieldQA-zh;
\textbf{(5-1 to 5-4)} denote HotpotQA, 2WikiMultihopQA, MuSiQue, and DuReader;
\textbf{(6-1 to 6-2)} denote LCC and RepoBench-P.}
\label{tab:Performance on LongBench}
\end{table*}

\begin{table*}[t]
\centering
\resizebox{0.95\linewidth}{!}{%
\begin{tabular}{c|cc|ccccccc|c}
\hline
Calibration Dataset & WikiText-2 & PTB & Openb. & ARC\_e & WinoG. & HeSW & ARC\_c & PIQA & MathQA & Average \\ \hline
WikiText-2 & 37.86 & 63.97 & 21.6 & 59.85 & 70.48 & 44.21 & 37.12 & 68.66 & 27.94 & 47.12 \\
C4 & 40.54 & 71.42 & 24 & 57.91 & 67.72 & 42.11 & 36.69 & 67.25 & 27.5 & 46.17 \\ \hline
\end{tabular}
}
\caption{Zero-shot performance of LLaMA3.1-8B-Instruct compressed by GRASP under 20\% compression using 512 samples from WikiText-2 and C4 as calibration datasets.}
\label{tab:Robustness of GRASP towards different representation evaluation dataset}
\end{table*}

\begin{table*}[t]
\centering
\resizebox{0.95\linewidth}{!}{%
\begin{tabular}{c|cc|ccccccc|c}
\hline
Calibration Dataset & WikiText-2 & PTB & Openb. & ARC\_e & WinoG. & HeSW & ARC\_c & PIQA & MathQA & Average \\ \hline
WikiText-2-64 & 46.5 & 86.51 & 22.6 & 59.97 & 69.3 & 44.36 & 37.29 & 68.5 & 27.37 & 47.06 \\
WikiText-2-128 & 39.91 & 76.41 & 22.8 & 60.23 & 69.93 & 44.24 & 35.92 & 67.9 & 27.5 & 46.93 \\
WikiText-2-256 & 38.73 & 79.13 & 21.8 & 59.89 & 70.24 & 44.23 & 36.26 & 67.19 & 27.07 & 46.67 \\
WikiText-2-512 & 37.86 & 63.97 & 21.6 & 59.85 & 70.48 & 44.21 & 37.12 & 68.66 & 27.94 & 47.12 \\ \hline
\end{tabular}
}
\caption{Zero-shot performance of LLaMA3.1-8B-Instruct compressed by GRASP under 20\% compression with varying calibration data sizes (64, 128, 256, 512) from WikiText-2.}
\label{tab:Robustness of GRASP towards different number of calibration data}
\end{table*}

\begin{table*}[t]
\centering
\resizebox{0.95\linewidth}{!}{%
    \begin{tabular}{c|ccccccc|c}
    \toprule[1.2pt]
    Method & Openb. & ARC\_e & ARC\_c & WinoG. & HeSW  & PIQA  & MathQA & Average \\
    \midrule
    \rowcolor[rgb]{ .91,  .91,  .91} Dense & 32.12 & 72.69 & 39.87 & 67.33 & 57.16 & 78.46 & 26.02 & 53.38 \\
    SparseGPT(2:4) & 24.40 & 64.10 & 30.80 & 66.61 & 43.09 & 70.73 & 24.25 & 46.28 \\
    wanda(2:4) & 24.60 & 62.75 & 30.80 & 62.67 & 41.38 & 70.29 & 23.62 & 45.16 \\
    GRASP & 24.80 & 56.02 & 32.68 & 66.14 & 42.57 & 70.74 & 23.92 & 45.27 \\
    GRASP* & 28.20 & 68.73 & 37.63 & 67.88 & 50.98 & 72.47 & 24.19 & 50.01 \\
    \bottomrule[1.2pt]
    \end{tabular}%
}
\caption{Comparison between GRASP and unstructured pruning methods under the 2:4 sparsity pattern (50\% sparsity). GRASP is applied at a 20\% structured compression ratio. GRASP* denotes the variant with light performance compensation}
\label{tab:Comparison_with_unstructured_pruning_methods}
\end{table*}

\end{document}